\documentclass[runningheads]{llncs}

\usepackage[mobile]{eccv}

\usepackage{eccvabbrv}

\usepackage{graphicx}
\usepackage{booktabs}

\usepackage{multirow,multicol} 
\usepackage{algorithm} 
\usepackage{algpseudocode} 
\usepackage{xcolor,colortbl} 
\DeclareUnicodeCharacter{221E}{\ensuremath{\infty}} 

\usepackage{wrapfig}
\usepackage{bbding}
\usepackage{pifont} 
\usepackage{latexsym}

\usepackage[accsupp]{axessibility}  

\usepackage[pagebackref,breaklinks,colorlinks,citecolor=eccvblue]{hyperref}

\usepackage{orcidlink}

\begin{document}

\title{Prompt-Driven Contrastive Learning for Transferable Adversarial Attacks}

\titlerunning{PDCL-Attack}

\author{Hunmin Yang\inst{1,2}\orcidlink{0009-0007-1625-2957} 
\and
Jongoh Jeong\inst{1}\orcidlink{0000-0002-5354-2693} 
\and
Kuk-Jin Yoon\inst{1}\orcidlink{0000-0002-1634-2756}
}

\authorrunning{H.~Yang et al.}

\institute{KAIST \and
ADD \\
\url{https://PDCL-Attack.github.io}
}

\maketitle

\begin{abstract}
Recent vision-language foundation models, such as CLIP, have demonstrated superior capabilities in learning representations that can be transferable across diverse range of downstream tasks and domains.
With the emergence of such powerful models, it has become crucial to effectively leverage their capabilities in tackling challenging vision tasks.
On the other hand, only a few works have focused on devising adversarial examples that transfer well to both unknown domains and model architectures.
In this paper, we propose a novel transfer attack method called \textbf{PDCL-Attack}, which leverages the CLIP model to enhance the transferability of adversarial perturbations generated by a generative model-based attack framework.
Specifically, we formulate an effective prompt-driven feature guidance by harnessing the semantic representation power of text, particularly from the ground-truth class labels of input images.
To the best of our knowledge, we are the first to introduce prompt learning to enhance the transferable generative attacks.
Extensive experiments conducted across various cross-domain and cross-model settings empirically validate our approach, demonstrating its superiority over state-of-the-art methods.
\keywords{Adversarial Attack \and Transferability \and Foundation Model \and Vision-Language Model \and Prompt Learning \and Contrastive Learning}
\end{abstract}

\begin{figure}
    \centering
    \includegraphics[width=0.93\linewidth]{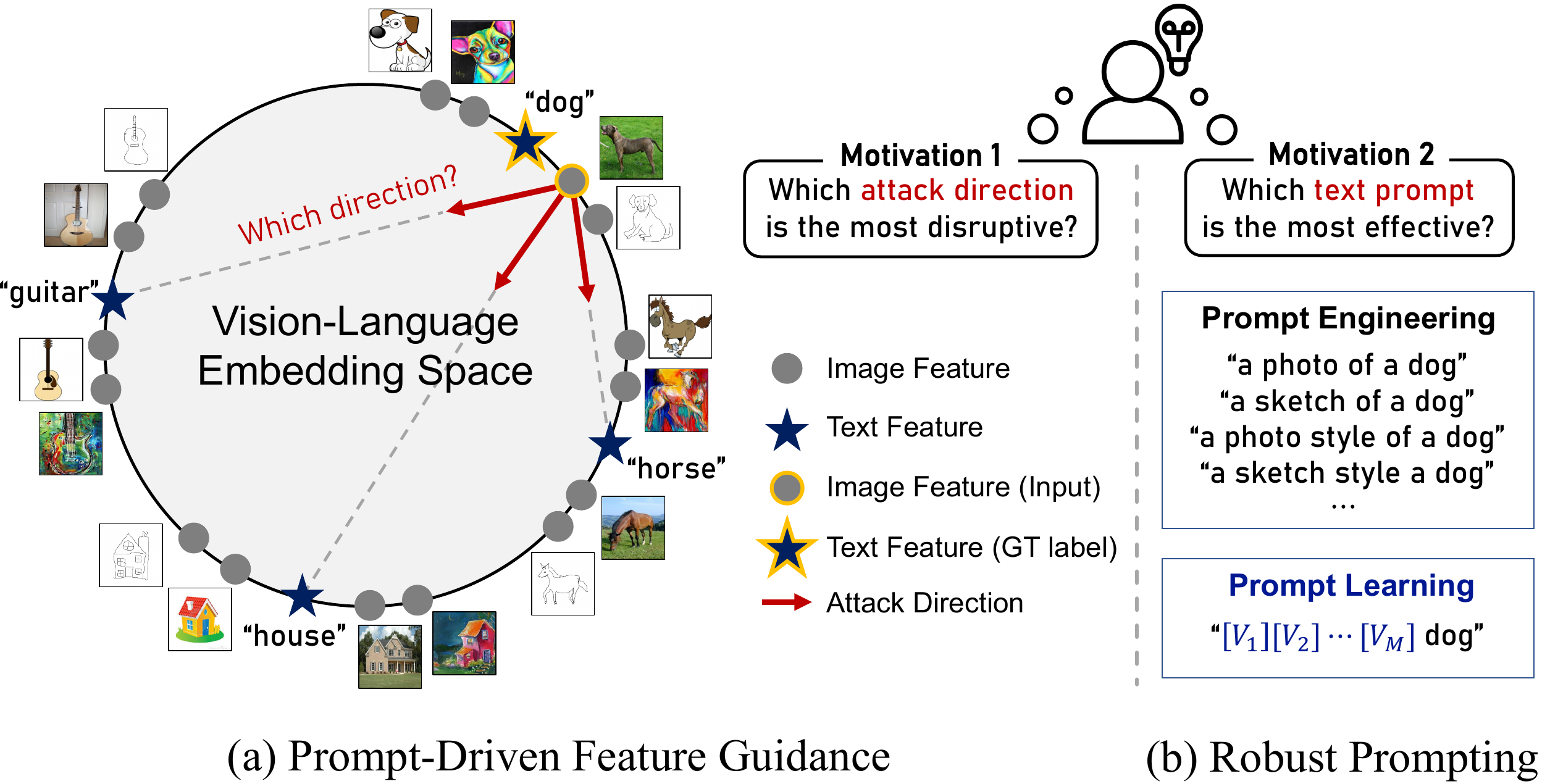}
    \caption{\textbf{Our motivation.}
    In a joint vision-language space, a single text can encapsulate core semantics that align with numerous images from diverse domains.
    Leveraging this principle, our approach utilizes representative prompt-driven text features to enhance the transferable adversarial attacks.
    On the adversary's side, two clear challenges arise: (a) Generating effective prompt-driven feature guidance, and (b) Identifying robust prompts which maximize the effectiveness.
    }
    \label{fig:motivation}
\end{figure}

\section{Introduction}
\label{sec:intro}
Regarding the training of deep neural networks, it is typically assumed that the training and testing data are independent and identically distributed.
This common principle may impair the generalization of trained models in situations involving domain shifts, which are frequently encountered in real-world deployments.
To address this issue, numerous approaches have been proposed in the fields of domain adaptation and generalization.
They fundamentally share a common objective: to extract domain-invariant semantic features that prove to be effective for target domains and downstream tasks.

Another related but distinct area of research delves into the transferability aspect of adversarial examples.
While these examples inherently possess somewhat transferable characteristics, na\"ively crafted adversarial examples are often limited in their transferability, particularly in strict black-box settings.
To address this challenge, researchers have explored transferable adversarial attacks with the aim of enhancing the transferability of adversarial examples.
Notably, recent generative model-based attacks~\cite{poursaeed2018generative, naseer2019cross, salzmann2021learning, naseer2021generating, zhang2022beyond, aich2022gama, yang2024facl} showcase superior adversarial transferability compared to iterative approaches~\cite{madry2017towards, croce2020reliable, lorenz2021detecting, DI, dr, ssp, long2022frequency, goodfellow2014explaining, kurakin2018adversarial, nguyen2015deep, carlini2017towards, szegedy2013intriguing, papernot2016limitations, dong2018boosting, chen2018ead}.
Furthermore, generative attacks possess a distinct advantage in inference time complexity over instance-specific iterative attacks, highlighting their practical utility in real-world scenarios.
GAP~\cite{poursaeed2018generative} initially established a generative attack framework, and CDA~\cite{naseer2019cross} introduced domain-agnostic generation of adversarial examples by employing relativistic cross-entropy loss.
LTP~\cite{salzmann2021learning} and BIA~\cite{zhang2022beyond} utilized the mid-layer features of surrogate models, which contain shared information across different architectures, making them effective for targeting various victim models.
FACL-Attack~\cite{yang2024facl} enhanced the transferability by employing a feature contrastive approach in the frequency domain.
Several other studies have employed vision-language models~\cite{aich2022gama} and object-centric features~\cite{aich2023leveraging} to acquire additional feature guidance for multi-object scene attacks.
As outlined in Table~\ref{tab:related_works}, attack strategies have evolved as adversaries leverage advanced open-source models, resulting in more effective feature guidance.

\begin{table}[!t]
\caption{
\textbf{Evolving attack strategies.}
This summarizes characteristics of generative model-based transfer attacks in a chronological order.
Adversaries are equipped with more powerful tools such as mid-layer features of surrogate model~\cite{salzmann2021learning, zhang2022beyond} and vision-language foundation model~\cite{aich2022gama}.
Our work further enhances the attack effectiveness by designing a novel prompt-driven attack loss (\textbf{Ours}\textsuperscript{\textdagger}) and employing a learnable prompt (\textbf{Ours}).
Please refer to Table~\ref{tab:ablation} for comparative evaluation results.
}
\setlength{\tabcolsep}{2.5pt}
\renewcommand{\arraystretch}{1.2}
\renewcommand{\aboverulesep}{0.1pt}
\renewcommand{\belowrulesep}{0.1pt}
\centering
\resizebox{\columnwidth}{!}{
\begin{tabular}{c|ccc|cccccc}
  \toprule

  \multirow{3}{*}{\textbf{Attack}} &
  \multicolumn{3}{c|}{\textbf{Surrogate Model}} &
  \multicolumn{6}{c}{\textbf{Vision-Language Foundation Model}} \\
  
  \cline{2-10} 
  
   &
  \multicolumn{2}{c|}{Output Logits} &
  Features &
  \multicolumn{2}{c|}{Image Features} &
  \multicolumn{2}{c|}{Text Features} &
  \multicolumn{2}{c}{Prompt Type} \\
  
  \cline{2-10} 
  
   &
  \multicolumn{1}{c|}{Absolute} &
  \multicolumn{1}{c|}{Relative} &
  Mid-layer &
  \multicolumn{1}{c|}{\;\;Clean\;\;} &
  \multicolumn{1}{c|}{\;\;\;Adv.\;\;\;} &
  \multicolumn{1}{c|}{\;\;\;\;GT\;\;\;\;} &
  \multicolumn{1}{c|}{\;\;\;Adv.\;\;\;} &
  \multicolumn{1}{c|}{\;Heuristic\;} &
  Learnable \\
  
  \hline
  \hline

  GAP~\cite{poursaeed2018generative} &
  \multicolumn{1}{c|}{\checkmark} &
  \multicolumn{1}{c|}{--} &
  -- &
  \multicolumn{1}{c|}{--} &
  \multicolumn{1}{c|}{--} &
  \multicolumn{1}{c|}{--} &
  \multicolumn{1}{c|}{--} &
  \multicolumn{1}{c|}{--} &
  -- \\
  
  CDA~\cite{naseer2019cross} &
  \multicolumn{1}{c|}{\checkmark} &
  \multicolumn{1}{c|}{\checkmark} &
  -- &
  \multicolumn{1}{c|}{--} &
  \multicolumn{1}{c|}{--} &
  \multicolumn{1}{c|}{--} &
  \multicolumn{1}{c|}{--} &
  \multicolumn{1}{c|}{--} &
  -- \\
  
  LTP~\cite{salzmann2021learning} &
  \multicolumn{1}{c|}{--} &
  \multicolumn{1}{c|}{--} &
  \checkmark &
  \multicolumn{1}{c|}{--} &
  \multicolumn{1}{c|}{--} &
  \multicolumn{1}{c|}{--} &
  \multicolumn{1}{c|}{--} &
  \multicolumn{1}{c|}{--} &
  -- \\

  BIA~\cite{zhang2022beyond} &
  \multicolumn{1}{c|}{--} &
  \multicolumn{1}{c|}{--} &
  \checkmark &
  \multicolumn{1}{c|}{--} &
  \multicolumn{1}{c|}{--} &
  \multicolumn{1}{c|}{--} &
  \multicolumn{1}{c|}{--} &
  \multicolumn{1}{c|}{--} &
  -- \\

  GAMA~\cite{aich2022gama} &
  \multicolumn{1}{c|}{--} &
  \multicolumn{1}{c|}{--} &
  \checkmark &
  \multicolumn{1}{c|}{\checkmark} &
  \multicolumn{1}{c|}{--} &
  \multicolumn{1}{c|}{--} &
  \multicolumn{1}{c|}{\checkmark} &
  \multicolumn{1}{c|}{\checkmark} &
  -- \\
  
  \hline
  
  \rowcolor{gray!9.0}\textbf{Ours}\textsuperscript{\textdagger} &
  \multicolumn{1}{c|}{--} &
  \multicolumn{1}{c|}{--} &
  \checkmark &
  \multicolumn{1}{c|}{\checkmark} &
  \multicolumn{1}{c|}{\checkmark} &
  \multicolumn{1}{c|}{\checkmark} &
  \multicolumn{1}{c|}{\checkmark} &
  \multicolumn{1}{c|}{\checkmark} &
  -- \\
  
  \rowcolor{gray!9.0}\textbf{Ours}&
  \multicolumn{1}{c|}{--} &
  \multicolumn{1}{c|}{--} &
  \checkmark &
  \multicolumn{1}{c|}{\checkmark} &
  \multicolumn{1}{c|}{\checkmark} &
  \multicolumn{1}{c|}{\checkmark} &
  \multicolumn{1}{c|}{\checkmark} &
  \multicolumn{1}{c|}{--} &
  \checkmark \\
  
  \bottomrule
\end{tabular}}
\label{tab:related_works}
\end{table}

In pursuit of acquiring highly generalizable representations, vision-language foundation models~\cite{CLIP, yao2021filip, jia2021scaling, yang2022unified, yang2022vision, you2022learning} 
have emerged.
These models have undergone large-scale training on vast amounts of web-scale data, yielding versatile features that serve as powerful priors for various downstream tasks.
In this context, we posited that generative attacks could also harness the capabilities of these potent models as illustrated in Fig.~\ref{fig:motivation}.
Our work draws inspiration from recent studies~\cite{LADS, PODA, PromptStyler, RISE} that employ CLIP~\cite{CLIP} as a tool for text-driven feature manipulation and domain extension.
These works have validated the superior representation power of CLIP, emphasizing a key insight that text can function as a prototype for various styles of images within a joint vision-language space.
PromptStyler~\cite{PromptStyler} further employs prompt learning~\cite{CoOp, CoCoOp} to improve the prompt-driven domain generalization performance.

Building upon this insight, we introduce a novel generative attack method that leverages  CLIP~\cite{CLIP} and prompt learning~\cite{CoOp} to effectively train the perturbation generator.
Our method, dubbed \textbf{PDCL-Attack}, integrates both pre-trained CLIP image and text encoders into the generative attack framework, further enhancing the effectiveness by employing a learnable prompt.
Our work is differentiated from GAMA~\cite{aich2022gama} in that we have re-designed the contrastive loss by separating the heterogeneous surrogate and CLIP features, and employing text features extracted from the ground-truth (GT) class labels of input images.
Our method incorporates perturbed image features from CLIP, which can provide effective loss gradients for training the robust generator.
To fully unleash the power of prompt-driven features, we pre-train the context words of the prompt to enhance the robustness of CLIP to distribution shifts.
In comparison to GAMA~\cite{aich2022gama}, our method with enhanced attack loss design (\textbf{Ours}\textsuperscript{\textdagger}, $1.87\%p\uparrow$) and incorporating prompt learning (\textbf{Ours}, $4.65\%p\uparrow$) demonstrates significant improvements.
Our contributions are summarized as follows:
\begin{itemize}
    \item[$\bullet$] We propose a novel prompt-driven contrastive learning (PDCL) method to enhance the training of the perturbation generator by leveraging the semantic representation power of text features from class labels of input images.
    \item[$\bullet$] This work is the first attempt to introduce prompt learning with a vision-language model to enhance the generative model-based transfer attacks.
    \item[$\bullet$] PDCL-Attack achieves state-of-the-art attack transferability across various domains and model architectures, surpassing previous approaches.
\end{itemize}

\section{Related Work}
\label{sec:related_work}
\noindent\textbf{Generative model-based attack.}\quad
Generative attacks~\cite{poursaeed2018generative, naseer2019cross, naseer2021generating, salzmann2021learning, zhang2022beyond, aich2022gama, aich2023leveraging, yang2024facl} leverage adversarial training~\cite{goodfellow2020generative}, using a pre-trained surrogate model as a discriminator to generate adversarial perturbations effective across entire data distributions.
This approach is computationally efficient and advantageous, as it can simultaneously generate diverse forms of perturbations across multiple images.
CDA~\cite{naseer2019cross} seeks to enhance the training of the generator by incorporating both the cross-entropy (CE) and the relativistic CE loss.
The work initially introduced domain-agnostic perturbations as well as model-agnostic ones.
LTP~\cite{salzmann2021learning} and BIA~\cite{zhang2022beyond} utilize features extracted from the mid-level layers of the surrogate model, which have been examined to contain a higher degree of shared information among various model architectures.
FACL-Attack~\cite{yang2024facl} exploits frequency domain manipulations to boost the attack transferability.
Several studies have utilized vision-language models~\cite{aich2022gama} and object-centric features~\cite{aich2023leveraging} to acquire effective features for attacking multi-object scenes.
Our work further enhances the training by employing a vision-language model and prompt learning.

\noindent\textbf{Vision-language foundation model.}\quad
Recent advancements in large vision-language models (VLM)~\cite{CLIP, yao2021filip, jia2021scaling, yang2022unified, yang2022vision, you2022learning} have demonstrated superior capabilities in learning generic representations.
This significant progress has been made possible by leveraging enormous web-scale training datasets.
In particular, Contrastive Language Image Pre-training (CLIP)~\cite{CLIP} employs 400 million image-text pairs for contrastive pre-training of image and text encoders within a joint vision-language embedding space.
The zero-shot CLIP model of ViT-L/14 exhibits impressive image classification performance, achieving a top-1 accuracy of 76.2\% on ImageNet-1K~\cite{imagenet}, on par with that of a fully-supervised ResNet101~\cite{res152}.
CLIP demonstrates superior domain generalization capability compared to supervised trained models.
For ImageNet-Sketch~\cite{imagenet-sketch}, zero-shot CLIP achieves 60.2\%, while the fully-supervised ResNet101 only attains 25.2\%.
Furthermore, context optimization~\cite{CoOp, CoCoOp} has further improved the few-shot performance of the employed CLIP model.
Interestingly, this trained context enhances CLIP's robustness to distribution shifts, as highlighted in~\cite{CoOp}.
Inspired by this line of research, our work incorporates CLIP and prompt learning into the generative attack framework to enable more effective transfer attacks.

\noindent\textbf{Text-driven feature guidance.}\quad
As both image and text features are mapped into a joint vision-language embedding space as in CLIP~\cite{CLIP}, various vision tasks focused on generalization or adaptation have benefited from text-driven feature guidance.
LADS~\cite{LADS} introduced a method for extending the trained model to new domains based solely on language descriptions of distribution shifts.
PODA~\cite{PODA} introduced a straightforward feature augmentation method for zero-shot domain adaptation, guided by a single textual description of the target domain.
PromptStyler~\cite{PromptStyler} proposed a text-driven style generation method to enhance the domain generalization performance.
From the perspective of utilizing generic text features to guide model training, our work aligns with these efforts.
We aim to explore an effective text-driven feature contrastive method to enhance the training of the perturbation generator towards a more robust regime.

\begin{figure*}[!t]
    \centering
    \includegraphics[width=\linewidth]{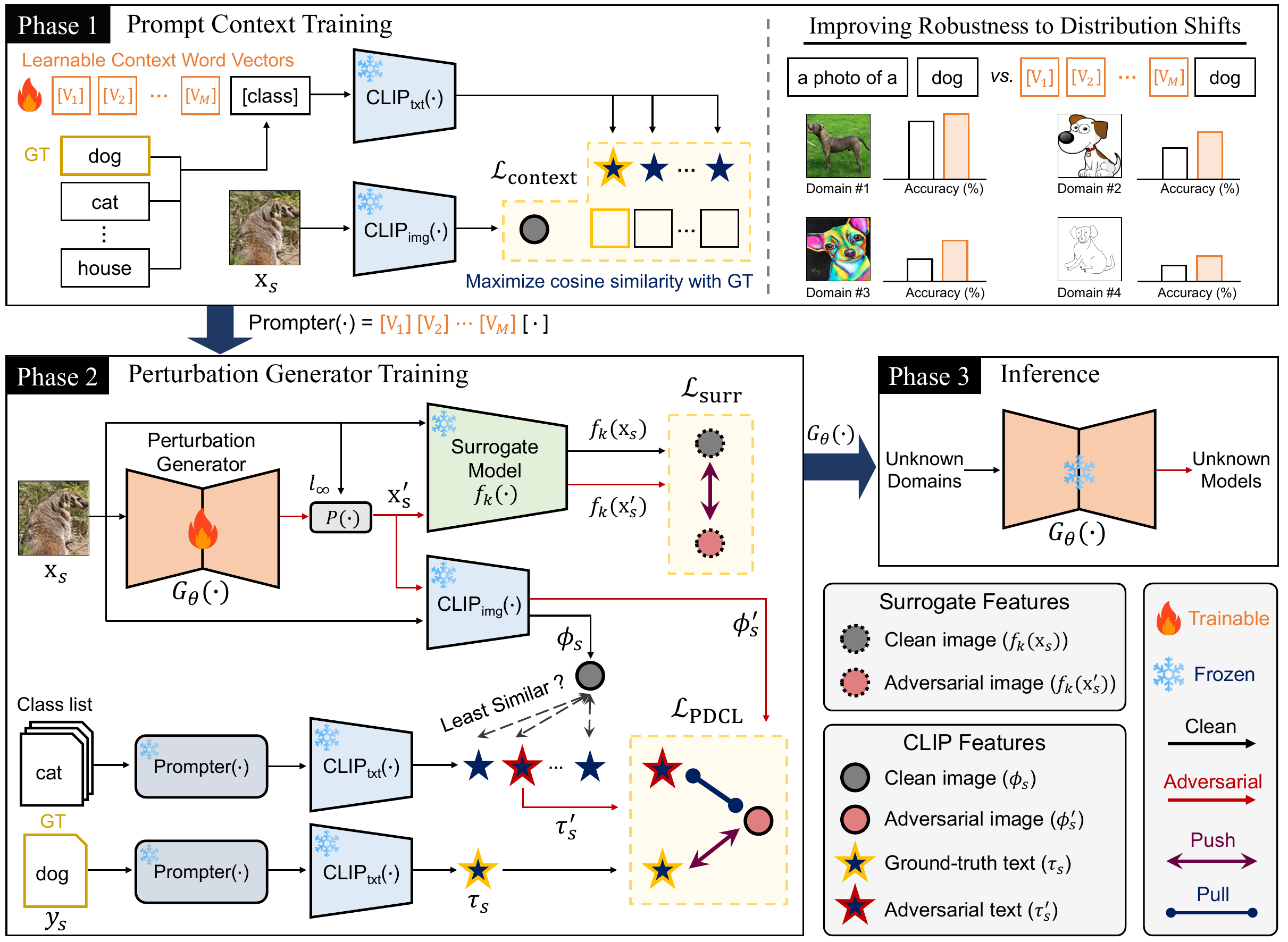}
    \caption{\textbf{Overview of PDCL-Attack}.
    For effective transfer attacks leveraging CLIP~\cite{CLIP}, our proposed pipeline consists of three serial stages; Phase 1 and 2 are the training stage, and Phase 3 is the inference stage.
    The goal of Phase 1 is to pre-train  $\texttt{Prompter}(\cdot)$, optimizing the context words $[\mathbf{V}_{1}] [\mathbf{V}_{2}] \cdots [\mathbf{V}_{M}]$ to yield generalizable text features in Phase 2.
    In Phase 1, only the learnable context word vectors are updated, while the weights of the CLIP image encoder $\mathrm{CLIP_{img}}(\cdot)$ and text encoder $\mathrm{CLIP_{txt}}(\cdot)$ remain fixed.
    In Phase 2, we train a generator model $G_{\theta}(\cdot)$ which crafts adversarial perturbations for encouraging a surrogate model $f_{k}(\cdot)$ to produce mispredictions for input images $\mathbf{x}_{s}$.
    The generator $G_{\theta}(\cdot)$ crafts the $\ell_{\infty}$-budget bounded adversarial image $\mathbf{x}'_{s}$ via a perturbation projector $P(\cdot)$.
    In Phase 3, we employ the trained generator $G_{\theta}(\cdot)$ to yield transferable adversarial examples on unknown domains and victim models.
    }
    \label{fig:overview}
\end{figure*}

\section{Proposed Approach: PDCL-Attack}
\label{sec:method}
\noindent\textbf{Problem definition.}\quad
Transfer attacks aim to craft adversarial examples which are transferable in black-box settings.
This task aims to create adversarial examples that are not just effective against a single surrogate model but also capable of deceiving other victim models, even those with different architectures or trained on different data distributions.
To that end, crafting transferable adversarial examples involves identifying vulnerabilities that are shared among diverse models and domains, effectively addressing the challenge of generalization.
Our goal is to devise an effective transfer attack method that can train a robust perturbation generator model, denoted as $G_{\theta}(\cdot)$, to generate adversarial examples which are transferable to arbitrary domains and model architectures.

\noindent\textbf{Overview of PDCL-Attack.}\quad
Overall framework of our proposed method is shown in Fig.~\ref{fig:overview}.
PDCL-Attack introduces a novel method for training a perturbation generator by leveraging CLIP~\cite{CLIP} and prompt learning~\cite{CoOp}.
Our attack pipeline consists of three sequential stages: Phases 1 and 2 constitute the training stage, whereas Phase 3 corresponds to the inference stage.
The objective of Phase 1 is to pre-train a $\texttt{Prompter}(\cdot)$ which can extract generalizable text features.
In Phase 1, learnable context word vectors of $\texttt{Prompter}(\cdot)$ are trained through few-shot learning by minimizing the cross-entropy loss on pairs of input images and their corresponding class labels.
In Phase 2, we train a perturbation generator $G_{\theta}(\cdot)$ which can craft adversarial perturbations that cause mispredictions of the surrogate model $f_{k}(\cdot)$.
By inputting the clean images $\mathbf{x}_{s}$, the generator $G_{\theta}(\cdot)$ crafts the $\ell_{\infty}$-bounded adversarial images $\mathbf{x}'_{s}$ via a perturbation projector $P(\cdot)$.
In Phase 3, we use the trained generator $G_{\theta}(\cdot)$ to craft transferable adversarial examples in black-box scenarios.

\subsection{Phase 1: Prompt Context Training}
The objective of Phase 1 is to train learnable context words used in $\texttt{Prompter}(\cdot)$, facilitating the extraction of generalizable text features in Phase 2.
To fully harness the capabilities of CLIP~\cite{CLIP} model, recent studies~\cite{CoOp, CoCoOp} have shown the effectiveness of leveraging context optimization.
These methods have demonstrated enhanced performance of the CLIP model across various distribution-shifting scenarios.
We adhere to the recent protocol outlined in~\cite{CoOp, CoCoOp} and integrate it to enhance our attack framework.
We utilize the same training dataset (\textit{i.e.}, ImageNet-1K~\cite{imagenet} which contains diverse distribution shifts) as in Phase 2.

\noindent\textbf{Optimal context training.}\quad
We model the context words used in $\texttt{Prompter}(\cdot)$ via learnable continuous vectors $[\mathbf{V}_{1}] [\mathbf{V}_{2}] \cdots [\mathbf{V}_{M}]$ as follows:
\begin{equation}
    \texttt{Prompter}(\cdot) = [\mathbf{V}_{1}] [\mathbf{V}_{2}] \cdots [\mathbf{V}_{M}] [\;\,\cdot\;\,],
\label{eq:prompter}
\end{equation}
where each $[\mathbf{V}_{m}]$ ($m \in \{1,...,M\}$) is a trainable 512-dimensional float vector with the same word embedding size of CLIP~\cite{CLIP}, and $M$ is the number of context words.
The training is exclusively focused on these word vectors, whereas the weights of the CLIP image encoder $\mathrm{CLIP_{img}}(\cdot)$ and text encoder $\mathrm{CLIP_{txt}}(\cdot)$ remain unchanged.
The training objective is to optimize $[\mathbf{V}_{1}] [\mathbf{V}_{2}] \cdots [\mathbf{V}_{M}]$ by minimizing the cross-entropy loss computed using pairs of input images and GT class labels through few-shot learning.
When the input images $\mathbf{x}_{s}$ are passed through $\mathrm{CLIP_{img}}(\cdot)$, the produced image features can be represented as $\phi_{s}$.
With $K$ class labels, we can obtain text features $\{\tau_{i}\}_{i=1}^{K}$ by feeding the class labels into $\mathrm{CLIP_{txt}}(\cdot)$.
Then, the prediction probability can be computed as:
\begin{equation}
    p(y=i|\mathbf{x}_{s}) = \frac{\mathrm{exp}(\mathrm{cos}(\tau_{i}, \phi_{s})/\lambda)}{\sum_{j=1}^{K}{\mathrm{exp}(\mathrm{cos}(\tau_{j}, \phi_{s})/\lambda)}},
\label{eq:probability}
\end{equation}
where $\lambda$ is the temperature parameter learned by CLIP~\cite{CLIP}, and $\mathrm{cos}(\cdot, \cdot)$ denotes the standard cosine similarity.

Using the prediction probability and a one-hot encoded label vector $l$, the context training loss can be computed as follows:
\begin{equation}
    \mathcal{L}_{\mathrm{context}} = -\sum_{j=1}^{K}{l_{j} \cdot \mathrm{log}\; p(y=j)},
\label{eq:loss_context}
\end{equation}
where the trained $\texttt{Prompter}(\cdot)$ along with $\mathrm{CLIP_{txt}}(\cdot)$ is utilized in Phase 2.

\begin{algorithm}[!t]
    \small
    \caption{Perturbation Generator Training (Phase 2)}
    \label{alg:PDCL-Attack}
    \begin{algorithmic}[1]
    \Require Source domain data distribution $\mathcal{X}_s$, perturbation generator $G_\theta(\cdot)$, 
    
    \;\;\, perturbation projector $P(\cdot)$, perturbation budget $\epsilon$, surrogate model $f_{k}(\cdot)$, 
    
    \;\;\, CLIP image encoder $\mathrm{CLIP_{img}}(\cdot)$, CLIP text encoder $\mathrm{CLIP_{txt}}(\cdot)$, $\texttt{Prompter}(\cdot)$
    \Ensure Freeze the pre-trained $f_{k}(\cdot)$, $\mathrm{CLIP_{img}}(\cdot)$, $\mathrm{CLIP_{txt}}(\cdot)$, $\texttt{Prompter}(\cdot)$

    \State Randomly initialize $G_\theta(\cdot)$
    
    \Repeat
    \State Randomly sample a mini-batch $\mathbf{x}_{s} \sim \mathcal{X}_{s}$ with batch size $n$
    \State Forward pass $\mathbf{x}_{s}$ through $G_{\theta}(\cdot)$ and generate unbounded 
    \Statex \;\;\;\;\, adversarial examples $\tilde{\mathbf{x}}'_{s}=G_{\theta}(\mathbf{x}_{s})$
    \State Bound the adversarial examples using $P(\cdot)$ such that:
    \begin{align*}
          \| P(\tilde{\mathbf{x}}'_{s}) - \mathbf{x}_s \|_{\infty} \le \epsilon
    \end{align*}
    \State Forward pass ${\mathbf{x}}'_s=P(\tilde{\mathbf{x}}'_s)$ and ${\mathbf{x}}_{s}$ through ${f}_{k}(\cdot)$
    \State Compute the surrogate model loss $\mathcal{L}_\mathrm{surr}$ using Eq.~(\ref{eq:loss_surr})
    \State Forward pass $\mathbf{x}'_{s}$ through $\mathrm{CLIP_{img}}(\cdot)$
    \State Forward pass class labels through $\mathrm{CLIP_{txt}}(\cdot)$ via $\texttt{Prompter}(\cdot)$
    \State Compute the prompt-driven contrastive loss $\mathcal{L}_{\mathrm{PDCL}}$ using Eq.~(\ref{eq:loss_PDCL})
    \State Compute the total loss $\mathcal{L}$ using Eq.~(\ref{eq:final_objective})
    \begin{align*}
          \mathcal{L} = \mathcal{L}_{\mathrm{surr}} + \mathcal{L}_{\mathrm{PDCL}}
    \end{align*}
    \State Backward pass and update $G_\theta(\cdot)$
    \Until{$G_\theta(\cdot)$ converges}
    \end{algorithmic}
\end{algorithm}

\subsection{Phase 2: Perturbation Generator Training}
This phase aims to train a perturbation generator model $G_{\theta}(\cdot)$ capable of crafting transferable perturbations as described in Algorithm~\ref{alg:PDCL-Attack}.
We randomly initialize the generator $G_\theta(\cdot)$.
Given a batch of clean images $\mathbf{x}_{s}$ with a batch size of $n$, $G_{\theta}(\cdot)$ generates unbounded adversarial examples. Subsequently, these are constrained by the perturbation projector $P(\cdot)$ within a predefined budget of $\epsilon$.

\noindent\textbf{Surrogate model loss} $\mathcal{L}_{\mathrm{surr}}$.\quad
The generated adversarial images $\mathbf{x}'_{s}$ and clean images $\mathbf{x}_{s}$ are passed through the surrogate model $f_{k}(\cdot)$, where we extract $k$-th mid-layer features.
We define the surrogate model loss $\mathcal{L}_{\mathrm{surr}}$ as follows:
\begin{equation}
    \mathcal{L}_{\mathrm{surr}} = \mathrm{cos} (f_{k}({\mathbf{x}}'_{s}), f_{k}(\mathbf{x}_{s})),
\label{eq:loss_surr}
\end{equation}
\noindent where $\mathrm{cos}(\cdot, \cdot)$ denotes the standard cosine similarity.

\noindent \textbf{Prompt-driven contrastive loss} $\mathcal{L}_{\mathrm{PDCL}}$.\quad
As CLIP~\cite{CLIP} employs contrastive learning to learn representations in a joint vision-language embedding space, we design a contrastive learning~\cite{hadsell2006dimensionality, chuang2020debiased} based attack loss for effectively harnessing CLIP features.
Our loss design differs from GAMA~\cite{aich2022gama} since we separate heterogeneous surrogate and CLIP features, and utilizes representative text features extracted from ground-truth (GT) class labels of input images.
Additionally, our method leverages CLIP's perturbed image features, which can provide effective loss gradients.
Given clean images $\mathbf{x}_{s}$ of batch size $n$, we first feed $\mathbf{x}_{s}$ into the CLIP image encoder $\mathrm{CLIP_{img}}(\cdot)$ for acquiring clean image features $\phi_{s}$.
We input $\mathbf{x}_{s}$ into $G_{\theta}(\cdot)$ followed by $P(\cdot)$ and craft adversarial images by ${\mathbf{x}}'_s=P(G_{\theta}(\mathbf{x}_{s}))$.
We then pass $\mathbf{x}_{s}'$ through $\mathrm{CLIP_{img}}(\cdot)$, and CLIP's adversarial image features $\phi'_{s}$ can be obtained as follows:
\begin{equation}
    \phi'_{s} = \mathrm{CLIP_{img}}(\mathbf{x}_{s}').
\label{eq:image_feature_adv}
\end{equation}

Since we utilize a large-scale ImageNet-1K~\cite{imagenet} dataset for training, we have corresponding GT class labels $y_{s}$ for each $\mathbf{x}_{s}$.
If there are $K$ classes in the training dataset (\textit{e.g.}, $K=1,000$ for ImageNet-1K), we can obtain the corresponding set of $K$ text features using the CLIP text encoder $\mathrm{CLIP_{txt}}(\cdot)$.
In contrast to GAMA~\cite{aich2022gama}, we pass $y_{s}$ as input to the pre-trained $\texttt{Prompter}(\cdot)$, which yields more effective text-driven prompt inputs for $\mathrm{CLIP_{txt}}(\cdot)$.
Then, text features $\tau_{s}$ extracted from the GT class labels of input images can be computed as follows:
\begin{equation}
    \tau_{s} = \mathrm{CLIP_{txt}}(\texttt{Prompter}(y_{s})).
\label{eq:text_feature_GT}
\end{equation}

With adversarial image features $\phi'_{s}$ as the anchor point, we employ adversarial text features $\tau'_{s}$ as the positives, which are computed by identifying the least similar text features compared to the input image features $\phi_{s}$.
Specifically, we compute a set of text features denoted as $\mathcal{T} = [\mathcal{T}_{1}, \mathcal{T}_{2}, \cdots, \mathcal{T}_{K}]$ using the class names of all $K$ classes, where each $\mathcal{T}_{i}$ corresponds to $\mathrm{CLIP_{txt}}(\texttt{Prompter}(y_{i}))$.
For each batch, we randomly select $n=16$ candidates from $\mathcal{T}$ for computational efficiency and identify the least similar candidate, using its label as $y'_{s}$.
We define the adversarial text features $\tau'_{s}$ as follows:
\begin{equation}
    \tau'_{s} = \mathrm{CLIP_{txt}}(\texttt{Prompter}(y'_{s})).
\label{eq:text_feature_adv}
\end{equation}

We use $\phi'_{s}$ from Eq.~(\ref{eq:image_feature_adv}) as the anchor, $\tau_{s}$ from Eq.~(\ref{eq:text_feature_GT}) as the negatives, and $\tau'_{s}$ from Eq.~(\ref{eq:text_feature_adv}) as the positives to constitute our contrastive loss.
Finally, our prompt-driven contrastive loss can be formulated as follows:
\begin{equation}
    \mathcal{L}_{\mathrm{PDCL}} = \|\phi'_{s} - \tau'_{s}\|_{2}^{2} \;+ \mathrm{max}(0, \; \alpha - \|\phi'_{s} - \tau_{s}\|_{2})^{2},
\label{eq:loss_PDCL}
\end{equation}
where $\alpha$ denotes the desired margin between $\phi'_{s}$ and $\tau_{s}$.
All the features are $\ell_{2}$-normalized before the loss calculation.
As a result, $\mathcal{L}_{\mathrm{PDCL}}$ facilitates more robust training of the generator $G_{\theta}(\cdot)$ by leveraging the prototypical semantic characteristics of text-driven features, boosted by the frozen robust $\texttt{Prompter}(\cdot)$.

\noindent\textbf{Total loss} $\mathcal{L}$.\quad
Our generator $G_{\theta}(\cdot)$ is trained by minimizing $\mathcal{L}$ as follows:
\begin{equation}
    \mathcal{L} = \mathcal{L}_{\mathrm{surr}} + \mathcal{L}_{\mathrm{PDCL}},
\label{eq:final_objective}
\end{equation}
where the total loss $\mathcal{L}$ facilitates $G_{\theta}(\cdot)$ to be trained towards more robust regime via effective prompt-driven feature guidance.

\subsection{Phase 3: Inference using the Frozen Generator}
Since our perturbation generator $G_{\theta}(\cdot)$ has been trained robustly, we freeze it for the final inference stage of crafting adversarial examples.
With the frozen $G_{\theta}(\cdot)$, we can generate adversarial examples by applying it to arbitrary image inputs, even if they belong to target data distributions different from the source domain.
We assess the performance of the trained $G_{\theta}(\cdot)$ in black-box scenarios across diverse domains and model architectures.
The generated adversarial examples are expected to exhibit superior cross-domain and cross-model transferability.

\section{Experiments}
\label{sec:experiments}
\subsection{Experimental Setting}
\noindent\textbf{Datasets and attack scenarios.}\quad
Since our goal is to generate adversarial examples which show high transferability across various domains and models, we carry out experiments in challenging black-box scenarios, namely \textit{cross-domain} and \textit{cross-model} settings.
Building upon a recent work~\cite{zhang2022beyond} that demonstrated remarkable attack transferability by employing a large-scale training dataset (\textit{i.e.}, ImageNet-1K~\cite{imagenet}), we also leverage ImageNet-1K to train our perturbation generator and prompter.
The efficacy of the trained generator is assessed by conducting evaluations on three additional datasets: CUB-201-2011~\cite{cub}, Stanford Cars~\cite{car}, and FGVC Aircraft~\cite{air}.
Specifically for the \textit{cross-domain} setting, we evaluate our method on unknown target domains (\textit{i.e.}, CUB-201-2011, Stanford Cars, FGVC Aircraft) and victim models distinct from both the source domain (\textit{i.e.}, ImageNet-1K) and the surrogate model.
For the \textit{cross-model} setting, we evaluate our method against black-box models with varying architectures, maintaining a white-box domain setup using ImageNet-1K.

\noindent\textbf{Surrogate and victim models.}\quad
The perturbation generator is trained on ImageNet-1K against a pre-trained surrogate model of VGG-16~\cite{vgg}.
For the \textit{cross-domain} setting, we employ fine-grained classification victim models trained by DCL framework~\cite{dcl}.
These models are based on three different backbones: ResNet50 (Res-50)~\cite{res152}, SENet154, and SE-ResNet101 (SE-Res101)~\cite{senet}.
For the \textit{cross-model} setting, we explore various different model architectures such as Res-50, ResNet152 (Res-152)~\cite{res152}, DenseNet121 (Dense-121), DenseNet169 (Dense-169)~\cite{densenet}, Inception-v3 (Inc-v3)~\cite{inc-v3}, MNasNet~\cite{tan2019mnasnet}, and ViT~\cite{dosovitskiy2020vit}.

\noindent\textbf{Implementation details.}\quad
We adhere closely to the implementations employed in recent generative model-based attacks~\cite{zhang2022beyond, aich2022gama} to ensure a fair comparison.
The mid-layer from which we extract features from the surrogate model (\textit{i.e.}, VGG-16~\cite{vgg}) is \textit{Maxpool.}3.
For the CLIP~\cite{CLIP} model, we use ViT-B/16~\cite{dosovitskiy2020vit} for the image encoder and Transformer~\cite{vaswani2017attention} for the text encoder, consistent with the configurations used in GAMA~\cite{aich2022gama}.
We train the perturbation generator using Adam optimizer~\cite{adam} with $\beta_{1}=0.5$ and $\beta_{2}=0.999$.
The learning rate is set to $0.0002$, and we use a batch size of $16$ for a single epoch training.
The perturbation budget for crafting the adversarial image is constrained to $\ell_{\infty} \leq 10$.
We use a contrastive loss margin of $\alpha=1$.
Regarding the prompt context training of Phase 1, we follow the standard protocol~\cite{CoOp, CLIP}.
Learnable word vectors are randomly initialized by a zero-mean Gaussian distribution with standard deviation of $0.02$.
We use a SGD optimizer with a learning rate of $0.002$, and train the word vectors with a maximum epoch of $50$ by 16-shot learning on ImageNet-1K~\cite{imagenet}.
The number of context words $M$ is set to $16$ and related analyses are shown in Table~\ref{tab:ablation}.
We also employ distribution-shifted versions of ImageNet (-V2~\cite{imagenet-v2}, -Sketch~\cite{imagenet-sketch}, -A~\cite{imagenet-a}, -R~\cite{imagenet-r}) to identify the robustness of the learned context words.
More details are provided in the Supplementary Material.

\noindent\textbf{Evaluation metric and competitors.}\quad
Our primary evaluation metric for assessing the attack effectiveness is the top-1 classification accuracy.
The competitors include state-of-the-art generative model-based attacks, such as GAP~\cite{poursaeed2018generative}, CDA~\cite{naseer2019cross}, LTP~\cite{salzmann2021learning}, BIA~\cite{zhang2022beyond}, and GAMA~\cite{aich2022gama}.
We train all the baselines~\cite{poursaeed2018generative, naseer2019cross, salzmann2021learning, zhang2022beyond, aich2022gama} on the same ImageNet-1K dataset for fair comparison.

\subsection{Main Results}
\noindent\textbf{Cross-domain evaluation results.}\quad
We compare our method with the state-of-the-art generative model-based attacks~\cite{poursaeed2018generative, naseer2019cross, salzmann2021learning, zhang2022beyond, aich2022gama} on various black-box domains with black-box models.
In the training stage, we utilize ImageNet-1K~\cite{imagenet} as the source domain to train a perturbation generator model against a pre-trained surrogate model of VGG-16~\cite{vgg}.
In the inference stage, we evaluate the trained perturbation generator on various unknown target domains, namely CUB-200-2011~\cite{cub}, Stanford Cars~\cite{car}, and FGVC Aircraft~\cite{air}, using different victim model architectures.
Specifically, we employ several fine-grained classification models that have been trained using the DCL framework~\cite{dcl}.
These victim models are based on three different backbones: Res-50~\cite{res152}, SENet154 and SE-ResNet101 (SE-Res101)~\cite{senet}.
We assess the effectiveness of our trained perturbation generator in this challenging black-box attack scenario.

As shown in Table~\ref{tab:cross_domain}, our method exhibits superior attack effectiveness with significant margins on most cross-domain benchmarks, which are also cross-model.
This highlights the robust and potent transferability of our crafted adversarial examples, enabled by prompt-driven feature guidance and prompt learning to maximize its effectiveness.
We conjecture that the remarkable generalization ability of PDCL-Attack might be attributed to the synergy between our two proposed methods: harnessing the superior representation power of CLIP's text features and improving it further with a prompt learning method to produce generalizable text features.
In essence, our approach indeed enhances the perturbation generator's capability to generalize across various black-box domains and state-of-the-art model architectures.

\begin{table*}[!t]
\caption{\textbf{Cross-domain evaluation results.} The perturbation generator is trained on ImageNet-1K~\cite{imagenet} with VGG-16~\cite{vgg} as the surrogate model and evaluated on black-box domains with models.
We compare the top-1 classification accuracy after attacks ($\downarrow$ is better) with the perturbation budget of $\ell_\infty \leq 10$.
\textbf{Best} and \underline{second best}.}
\setlength{\tabcolsep}{1.1pt}
\renewcommand{\arraystretch}{1.2}
\renewcommand{\aboverulesep}{0.1pt}
\renewcommand{\belowrulesep}{0.1pt}
\centering
\resizebox{\linewidth}{!}{
    \begin{tabular}{ccccccccccccc}
        \toprule
        \multirow{2}{*}{Method} & \multicolumn{3}{c}{CUB-200-2011} && \multicolumn{3}{c}{Stanford Cars} && \multicolumn{3}{c}{FGVC Aircraft} & \multirow{2}{*}{AVG.} \\
        \cmidrule{2-12}
         
        & Res-50 & SENet154 & SE-Res101 && Res-50 & SENet154 & SE-Res101 &&  Res-50 & SENet154 & SE-Res101 \\
        \midrule
        \midrule
        
        Clean & 87.33 & 86.81 & 86.59 && 94.25 & 93.35 & 92.96 && 92.14 & 92.05 & 91.84 & 90.81 \\
        \midrule
        
        GAP~\cite{poursaeed2018generative} & 68.85 & 74.11 & 72.73 && 85.64 & 84.34 & 87.84 && 81.40 & 81.88 & 76.90 & 79.30 \\
        
        CDA~\cite{naseer2019cross} & 69.69 & 62.51 & 71.00 && 75.94 & 72.45 & 84.64 && 71.53 & 58.33 & 63.39 &69.94 \\
        
        LTP~\cite{salzmann2021learning} & \underline{30.86} & \underline{52.50} & 62.86 && 34.54 & \underline{65.53} & 73.88 && \underline{15.90} & 60.37 & 52.75 & 49.91 \\
        
        BIA~\cite{zhang2022beyond} & 32.74 & 52.99 & 58.04 && 39.61 & 69.90 & 70.17 && 28.92 & 60.31 & \underline{46.92} & 51.07 \\

        GAMA~\cite{aich2022gama} & 34.47 &54.02 &	\underline{57.66} &&	\underline{30.16} &	69.80 &	\underline{63.82} &&	25.29 &	\underline{58.42} &	\textbf{43.41} & \underline{48.56} \\
        \midrule
        
        \rowcolor{gray!9.0}\textbf{Ours} & \textbf{22.97}  & 	\textbf{49.19}  & \textbf{54.92}  && 	\textbf{22.58}  & 	\textbf{64.95}  & \textbf{63.70}  && 	\textbf{15.81}  & 	\textbf{53.83}  & 47.25  & 	\textbf{43.91}  \\ 
        
        \bottomrule
    \end{tabular}
}
\label{tab:cross_domain}
\end{table*}

\noindent\textbf{Cross-model evaluation results.}\quad
While our method demonstrates the effectiveness in enhancing the attack transferability in the strict black-box scenario as shown in Table~\ref{tab:cross_domain}, we further conducted investigations in a controlled white-box domain scenario, specifically within ImageNet-1K~\cite{imagenet}.
We train a generator against a surrogate model of VGG-16~\cite{vgg}, and subsequently assess its effectiveness on victim models with various architectures such as ResNet50 (Res-50), ResNet152 (Res-152)~\cite{res152}, DenseNet121 (Dense-121), DenseNet169 (Dense-169)~\cite{densenet}, Inception-v3 (Inc-v3)~\cite{inc-v3}, MNasNet~\cite{tan2019mnasnet}, and ViT~\cite{dosovitskiy2020vit}.

As shown in Table~\ref{tab:cross_model}, our method mostly achieves competitive performance even in the white-box domain scenario.
We conjecture that our CLIP-driven guidance, coupled with optimized prompt context, can improve the training of the generator in crafting more resilient perturbations, ultimately showcasing enhanced generalization capabilities in unknown feature spaces of victim models.
It is noteworthy that incorporating the pre-trained CLIP ViT-B/16~\cite{dosovitskiy2020vit} image encoder alongside the surrogate model might be one of the factors which could enhance the ViT-based transferability.
Nonetheless, compared to GAMA~\cite{aich2022gama} which also utilizes the same CLIP encoder backbone, our method surpasses it particularly on ViT-based model transferability.
We posit that incorporating text feature guidance from GT labels and adversarial image-driven loss gradients further improves the training of the perturbation generator.

\begin{table*}[!t]
\caption{\textbf{Cross-model evaluation results.}
The perturbation generator is trained on ImageNet-1K~\cite{imagenet} with VGG-16~\cite{vgg} as the surrogate model and evaluated on black-box models.
We compare the top-1 classification accuracy after attacks ($\downarrow$ is better) with the perturbation budget of $\ell_\infty \leq 10$.
\textbf{Best} and \underline{second best}.}
\setlength{\tabcolsep}{2.5pt}
\renewcommand{\arraystretch}{1.2}
\renewcommand{\aboverulesep}{0.1pt}
\renewcommand{\belowrulesep}{0.1pt}
\centering
\resizebox{\linewidth}{!}{
    \begin{tabular}{cccccccccc}
        \toprule 
        Method &
        Res-50 &
        Res-152 &
        Dense-121 &
        Dense-169 &
        Inc-v3 &
        MNasNet &
        ViT-B/16 &
        ViT-L/16 &
        AVG. \\
        \midrule
        \midrule        
            Clean & 74.61 & 77.34 & 74.22 & 75.75 & 76.19 & 66.49 & 79.56 & 80.86 & 75.63 \\
             
            \midrule
    
            GAP~\cite{poursaeed2018generative} & 
            57.87 & 65.50 & 57.94 & 61.37 & 63.30 & 42.47 & 72.89 & 76.69 & 62.25 \\
              
            CDA~\cite{naseer2019cross} & 36.27 & 51.05 & 38.89 & 42.67 &  54.02 & 33.10 & 68.73 & 74.22 & 49.87 \\
            
            LTP~\cite{salzmann2021learning} & 
            \underline{21.70} & \underline{39.88} & \underline{23.42} & \textbf{25.46} & 41.27 & 45.28 & 72.44 & 76.75 & 43.28 \\
            
            BIA~\cite{zhang2022beyond} & 
            25.36 & 42.98 & 26.97 & 32.35 & 41.20 & 34.31 & \underline{67.05} & 73.23 & 42.93\\

            GAMA~\cite{aich2022gama} & 
            24.82 & 43.22 & 24.84 & 30.81 & \underline{35.10} & \textbf{27.96} & 67.33 & \underline{73.16} & \underline{40.91} \\
            \midrule

            \rowcolor{gray!9.0}\textbf{Ours} & \textbf{20.87} & \textbf{38.62} & \textbf{21.26} & \underline{29.01} & \textbf{32.99} & \underline{28.00} & \textbf{65.53} & \textbf{72.52} & \textbf{38.60} \\
             	
        \bottomrule
    \end{tabular}
}
\label{tab:cross_model}
\end{table*}
\begin{table}[!t]
\caption{\textbf{Ablation study on our proposed losses.}
With the same surrogate model loss $\mathcal{L}_{\mathrm{surr}}$, \textbf{Ours}\textsuperscript{\textdagger} using a hand-crafted prompt (\ie, ``a photo of a [class]'') outperforms $\mathcal{L}_{\mathrm{GAMA}}$~\cite{aich2022gama} by our improved CLIP-guidance loss $\mathcal{L}_{\mathrm{PDCL}}$.
\textbf{Ours} using a learned prompt (\ie, ``$[\mathbf{V}_{1}] [\mathbf{V}_{2}] \cdots [\mathbf{V}_{16}]$ [class]'') further enhances the attack effectiveness by our context training loss $\mathcal{L}_{\mathrm{context}}$.
\textbf{Best} and \underline{second best}.
}
\centering
\setlength{\tabcolsep}{4pt} 
\renewcommand{\arraystretch}{1.2}
\renewcommand{\aboverulesep}{0.1pt}
\renewcommand{\belowrulesep}{0.1pt}
\resizebox{0.95\linewidth}{!}{
\begin{tabular}{ccccccc}
\toprule

\multicolumn{1}{c|}{\multirow{1}{*}{Method}} &
\multicolumn{1}{c}{\multirow{1}{*}{$\mathcal{L}_{\mathrm{surr}}$}} &
\multicolumn{1}{c|}{\multirow{1}{*}{$\mathcal{L}_{\mathrm{GAMA}}$}} &
\multicolumn{1}{c}{\multirow{1}{*}{$\mathcal{L}_{\mathrm{PDCL}}$}} &
\multicolumn{1}{c|}{\multirow{1}{*}{$\mathcal{L}_{\mathrm{context}}$}} &
\multicolumn{1}{c}{\small{Cross-Domain}} & \multicolumn{1}{c}{\small{Cross-Model}} \\

\midrule
\midrule

\multicolumn{1}{c|}{Clean} & -- & \multicolumn{1}{c|}{--} & -- & \multicolumn{1}{c|}{--} & 90.85 & 75.63 \\

\midrule

\multicolumn{1}{c|}{BIA~\cite{zhang2022beyond}} & \checkmark &
\multicolumn{1}{c|}{--} & -- &\multicolumn{1}{c|}{--} & 51.07 & 42.93 \\

\multicolumn{1}{c|}{GAMA~\cite{aich2022gama}} & \checkmark & \multicolumn{1}{c|}{\checkmark} & -- &\multicolumn{1}{c|}{--} & 48.56 & 40.91 \\

\midrule

\rowcolor{gray!9.0}\multicolumn{1}{c|}{\textbf{Ours}\textsuperscript{\textdagger}} & \checkmark & \multicolumn{1}{c|}{--} & \checkmark & \multicolumn{1}{c|}{--} & \underline{46.69} & \underline{40.35} \\

\rowcolor{gray!9.0}\multicolumn{1}{c|}{\textbf{Ours}} & \checkmark & \multicolumn{1}{c|}{--} & \checkmark & \multicolumn{1}{c|}{\checkmark} & \textbf{43.91} & \textbf{38.60} \\

\bottomrule
\end{tabular}
}
\label{tab:ablation}
\end{table}

\subsection{More Analyses}
\noindent\textbf{Ablation study on our proposed losses.}\quad
\label{sec:ablation}
In transfer-based attack scenarios, the adversary leverages surrogate models to simulate potential unknown victim models and craft adversarial examples.
In this work, we additionally leverage a vision-language foundation model (\textit{i.e.}, CLIP~\cite{CLIP}) and formulate effective loss functions to enhance the training.
In Table~\ref{tab:ablation}, we evaluate the effect of our proposed losses used for improving the transferability of adversarial examples in both cross-domain and cross-model scenarios.
Remarkably, in \textbf{Ours}\textsuperscript{\textdagger}, our approach achieves state-of-the-art results even without employing prompt learning.
This highlights the effectiveness of our proposed CLIP-driven attack loss $\mathcal{L}_{\mathrm{PDCL}}$ compared to GAMA~\cite{aich2022gama}.
This demonstrates the effectiveness of the text-driven semantic feature guidance from GT labels and back-propagated loss gradients from adversarial CLIP image features.
Moreover, incorporating prompt learning $\mathcal{L}_{\mathrm{context}}$ into the framework further boosts the transferability as demonstrated in \textbf{Ours}.
When employing $\texttt{Prompter}(\cdot)$ trained with $\mathcal{L}_{\mathrm{context}}$, the text-driven guidance facilitated by $\mathcal{L}_{\mathrm{PDCL}}$ is strengthened by the extraction of more generalizable features.
Compared to the baselines, our method consistently improves both cross-domain and cross-model attack transferability.

\begin{table}[t]
    \caption{\textbf{Effect of learnable context words.}
    Learnable context words outperform hand-crafted heuristic ones, and increasing their capacity further improves the attack effectiveness.
    \textbf{Best} and \underline{second best}.
    }
    \centering
    \setlength{\tabcolsep}{4.5pt}
    \renewcommand{\arraystretch}{1.2}
    \renewcommand{\aboverulesep}{0.1pt}
    \renewcommand{\belowrulesep}{0.1pt}
    \resizebox{0.85\linewidth}{!}{
    \begin{tabular}{c|c|c|c}
         \toprule
         
         Type & \# of words & Text Prompt & Accuracy~($\downarrow$) \\
         
         \midrule
         \midrule
         
         \multirow{5}{*}{Heuristic} & \multirow{2}{*}{$M=4$} & ``a photo of a [class]'' & 46.69 \\
         
          &  & ``a sketch of a [class]'' & 47.02 \\
          
         \cmidrule{2-4}
         
          & \multirow{3}{*}{$M=5$} & ``a photo style of a [class]'' & 46.14 \\
          
          &  & ``a sketch style of a [class]'' & 47.70 \\
          
          &  & ``a $[\mathbf{V}_{\mathrm{rand}}]$ style of a [class]'' & 47.81 \\
          
         \midrule
         
         \rowcolor{gray!9.0} & $M=4$ & ``$[\mathbf{V}_{1}] [\mathbf{V}_{2}][\mathbf{V}_{3}] [\mathbf{V}_{4}] $ [class]'' & \underline{45.44} \\
         
         \rowcolor{gray!9.0} \multirow{-2}{*}{Learnable} & $M=16$ & ``$[\mathbf{V}_{1}] [\mathbf{V}_{2}] \cdots [\mathbf{V}_{16}]$ [class]'' & \textbf{43.91} \\
         
         \bottomrule
    \end{tabular}
    }
    \label{tab:prompt_engineering}
\end{table}

\noindent\textbf{Effect of learnable context words.}\quad
In Table~\ref{tab:prompt_engineering}, we compare our prompt learning method with various hand-crafted text prompts in a cross-domain attack scenario.
Using domain-specific heuristic prompts, such as ``a [domain] of a [class]'' or ``a [domain] style of a [class]'', CLIP's textual guidance might fall into sub-optimal regime due to the domain-specific overfitting.
Considering this, we investigated a domain randomization approach by randomly initializing a word vector, \textit{i.e.}, ``a $[\mathbf{V}_{\mathrm{rand}}]$ style of a [class]''.
In this trial, we randomly initialize $[\mathbf{V}_{\mathrm{rand}}]$ for each training iteration.
However, the effectiveness is comparable to that of the na\"ive heuristic prompts, indicating the necessity for a meticulously crafted prompt to attain better performance.
We conjecture that our superior results might be attributed to the enhanced representational power of generalizable text features attained through learned context words.
This also aligns with recent findings~\cite{CoOp, CoCoOp} that prompt learning not only enhances the downstream task performance, it can even improve the robustness of the trained model in distribution-shifted scenarios.
Remarkably, increasing the capacity of learnable context words ($M=16$) further enhances the attack effectiveness.

\begin{table}[t]
    \caption{\textbf{Cross-domain attack effectiveness w.r.t. perturbation budget.}
    Given the same test-time perturbation budget, our method consistently demonstrates superior attack effectiveness.
    \textbf{Best} and \underline{second best}.
    }
    \centering
    \setlength{\tabcolsep}{5pt}
    \renewcommand{\arraystretch}{1.2}
    \renewcommand{\aboverulesep}{0.1pt}
    \renewcommand{\belowrulesep}{0.1pt}
    \resizebox{0.65\columnwidth}{!}{
    \begin{tabular}{c|ccccc}
         \toprule
         $\ell_{\infty} \leq$ & 6 & 7 & 8 & 9 & 10 \\
         \midrule
         \midrule
         
         BIA~\cite{zhang2022beyond} & 78.76 & 72.32  & 65.03 & 57.75 & 51.07 \\
         GAMA~\cite{aich2022gama} & \underline{75.23} & \underline{68.41} & \underline{61.52} & \underline{54.66} & \underline{48.56} \\
         \midrule

         \rowcolor{gray!9.0}\textbf{Ours} & \textbf{72.07} & \textbf{64.67} & \textbf{57.04} & \textbf{50.23} & \textbf{43.91} \\
         \bottomrule
    \end{tabular}
    }
    \label{tab:perturbation_budget}
\end{table}

\begin{figure}[t]
    \centering
    \includegraphics[width=0.99\linewidth]{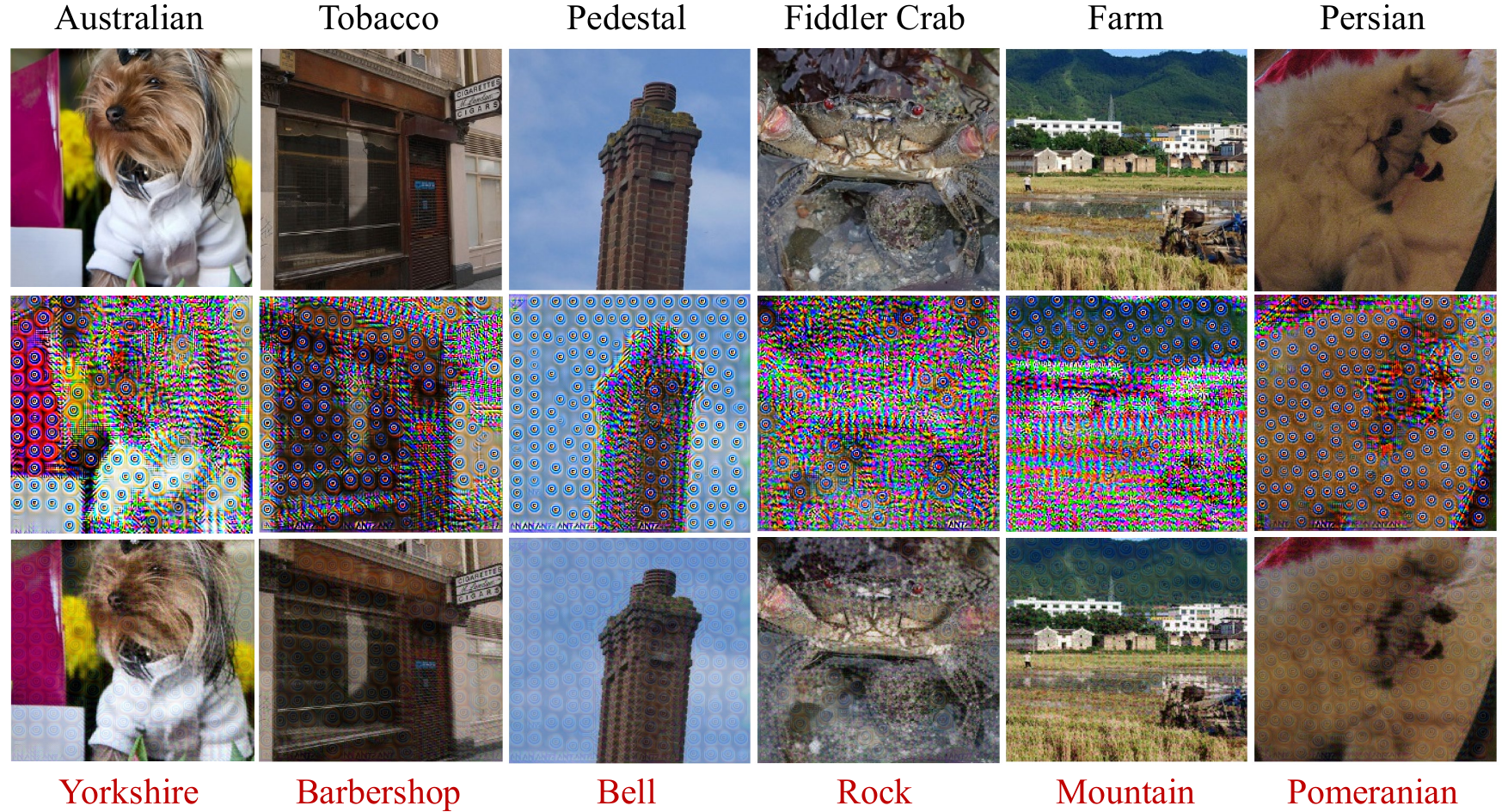}
    \caption{\textbf{Qualitative results.}
    PDCL-Attack successfully fools the classifier, causing it to predict the clean image labels (in black) as the mispredicted class labels shown at the bottom (in red).
    From top to bottom: clean images, unbounded adversarial images, and bounded ($\ell_\infty \leq 10$) adversarial images which are \textit{actual inputs} to the classifier.
    }
    \label{fig:qualitative}
\end{figure}

\noindent\textbf{Attack effectiveness and perceptibility.}\quad
While our work primarily focuses on crafting more effective perturbations, it is also crucial to carefully examine the image quality of the adversarial examples for real-world deployment.
Therefore, we investigate how the perturbation budget affects the transferability of cross-domain attacks in Table~\ref{tab:perturbation_budget}.
We compare the top-1 classification accuracy after attacks using the generator trained with the perturbation budget of $\ell_{\infty} \leq 10$.
Across each test-time perturbation budget, our method consistently demonstrates superior attack performance.
In other words, ours can achieve higher attack transferability with lower perturbation power and better image quality, which are significant advantages in real-world scenarios.
PDCL-Attack can generate effective and high-quality adversarial images as shown in Fig.~\ref{fig:qualitative}.

\section{Limitations and Broader Impacts}
\label{sec:discussion}
\noindent\textbf{Limitations.}\quad
Since our method employs a vision-language foundation model, it entails several limitations.
Firstly, the attack effectiveness depends on the quality of the pre-trained vision-language model.
This could be alleviated with the progress of vision-language foundation models.
Secondly, our approach leveraging CLIP~\cite{CLIP} incurs higher training computational costs.
For instance, our method takes longer training time than BIA~\cite{zhang2022beyond} ($\sim$33 vs. 12 hours).
Thirdly, our method necessitates the availability of ground-truth class labels for text-driven feature guidance.
Lastly, we limited our evaluation to an image classification task.
We posit that CLIP's generalizable text features could also offer benefits across various tasks, and we defer this exploration to future work.

\noindent\textbf{Broader impacts.}\quad
Our research community is facing remarkable progress regarding large language models (LLMs) and multi-modal foundation models.
It is crucial to note that adversaries also have access to these powerful open-source models, and potential threats are escalating with their advancement.
This work aims to raise awareness of such risks and encourage research into developing methodologies for training more robust models.

\section{Conclusion}
\label{sec:conclusion}
In this paper, we have introduced a novel generator-based transfer attack method that harnesses the capabilities of a vision-language foundation model and prompt learning.
We design an effective attack loss by incorporating image and text features extracted from the CLIP model and integrating these with a surrogate model-based loss.
We find that text feature guidance from ground-truth labels and adversarial image-driven loss gradients enhance the training.
Extensive evaluation results validate the effectiveness of our approach in black-box scenarios with unknown distribution shifts and variations in model architectures.

\section*{Acknowledgements}
This work was partially supported by the Agency for Defense Development grant funded by the Korean Government.
We thank Junhyeong Cho for his insightful discussions and valuable comments.

%
%

\clearpage
\title{Prompt-Driven Contrastive Learning for Transferable Adversarial Attacks\\
\textmd{--- Supplementary Material ---}}

\titlerunning{PDCL-Attack}

\author{Hunmin Yang\inst{1,2}\orcidlink{0009-0007-1625-2957} \and
Jongoh Jeong\inst{1}\orcidlink{0000-0002-5354-2693} \and
Kuk-Jin Yoon\inst{1}\orcidlink{0000-0002-1634-2756}
}

\authorrunning{H.~Yang et al.}

\institute{KAIST \and
ADD \\
\url{https://PDCL-Attack.github.io}
}

\maketitle
In this supplementary material, we provide additional details that are not included in the paper due to space limitations.
This includes the preliminaries (Sec.~\ref{sec:preliminaries}), implementation details (Sec.~\ref{sec:supp_implementation_details}), more quantitative experimental results (Sec.~\ref{sec:supp_quantitative}), and more qualitative experimental results (Sec.~\ref{sec:supp_qualitative}), respectively.

\setcounter{figure}{0}
\setcounter{table}{0}
\setcounter{algorithm}{0}
\setcounter{equation}{0}
\renewcommand{\thefigure}{A\arabic{figure}}
\renewcommand{\thetable}{A\arabic{table}}
\renewcommand{\thealgorithm}{A\arabic{algorithm}}
\renewcommand{\theequation}{A\arabic{equation}}
\renewcommand{\thesection}{A}

\section{Preliminaries}
\label{sec:preliminaries}
\subsection{Generative Model-based Adversarial Attack}
In previous years, the standard attack protocol has been instance-based iterative methods~\cite{madry2017towards, croce2020reliable, lorenz2021detecting, DI, dr, ssp, long2022frequency, goodfellow2014explaining, kurakin2018adversarial, nguyen2015deep, carlini2017towards, szegedy2013intriguing, papernot2016limitations, dong2018boosting, chen2018ead}, favored for their simplicity and effectiveness.
However, the iterative approach often encounters limitations due to its inefficient time complexity and overfitting to the training data and model.
In this context, generative attacks~\cite{poursaeed2018generative, naseer2019cross, naseer2021generating, salzmann2021learning, zhang2022beyond, yang2024facl} have garnered attention, demonstrating their high transferability across various domains and models.
This is attributed to the fact that generative attacks utilize entire data distributions through adversarial training~\cite{goodfellow2020generative}, rather than fitting individual image samples.

\noindent\textbf{Perturbation generator.}\quad 
Given a batch of clean image samples $\mathbf{x}$, the perturbation generator $G_\theta(\cdot)$ generates unbounded adversarial examples $\tilde{\mathbf{x}}'$.
These unbounded samples are constrained within a predefined budget $\epsilon$ by the perturbation projector $P(\cdot)$, ensuring that $\| P(\tilde{\mathbf{x}}') - \mathbf{x} \|_{\infty} \le \epsilon$.
The resulting bounded adversarial examples $\mathbf{x}'$ are designed to perturb the predictions of unknown victim models.
Our setting is established within a rigorous black-box scenario, in which both the target domain distributions and model architectures vary.

\noindent\textbf{Surrogate model.}\quad
In transfer-based attack scenarios, the adversary leverages surrogate models to simulate potential unknown victim models and craft adversarial examples.
The perturbation generator $G_\theta(\cdot)$ is trained by perturbing the predictions of the surrogate model and back-propagating the loss gradients.
Recent works~\cite{salzmann2021learning, zhang2022beyond, wu2020boosting, aich2022gama, aich2023leveraging} utilize mid-layer features $f_{k}(\cdot)$ from the surrogate model, which encapsulate shared attributes among various model architectures.
Building upon these approaches, we utilize a pre-trained CLIP~\cite{CLIP} model as an additional guidance model that has undergone large-scale training.

\subsection{Vision-Language Foundation Model}
Among the plethora of large-scale vision-language models (VLMs)~\cite{CLIP, yao2021filip, jia2021scaling, yang2022unified, yang2022vision, you2022learning, yang2024facl}, we choose CLIP~\cite{CLIP} as our employed pre-trained vision-language model.
Although we use CLIP in this work, our method can also be applicable to other CLIP-like VLMs as well.
In the case of input batches containing image-text pairs, CLIP jointly train an image encoder and a text encoder using cosine similarity scores derived from each image-text pair.
Specifically, CLIP employs 400 million image-text pairs for contrastive training of image and text encoders within a shared vision-language embedding space.
CLIP exhibits exceptional capabilities in learning generic visual representations, facilitating seamless zero-shot transfer to diverse downstream vision tasks and domains.

\noindent\textbf{Joint vision-language training.}\quad
In datasets consisting of $N \times N$ image-text pairs, the $N$ matched pairs are considered positive pairs, while the remaining $N^2 - N$ pairs are considered negative pairs.
CLIP is trained by maximizing the cosine similarities between image and text features from the positive pairs, while minimizing the similarities between such features from the negative pairs.

\noindent\textbf{Image encoder.}\quad
CLIP utilizes either ViT~\cite{dosovitskiy2020vit} or ResNet~\cite{res152} architecture as the image encoder $\mathrm{CLIP_{img}}(\cdot)$.
$\mathrm{CLIP_{img}}(\cdot)$ processes input images to extract their corresponding image features.
Then, the extracted image features are $\ell_{2}$-normalized to lie on a hypersphere within a joint vision-language embedding space.
We employ the pre-trained $\mathrm{CLIP_{img}}(\cdot)$ and keep it frozen.

\noindent\textbf{Text encoder.}\quad
CLIP utilizes Transformer~\cite{vaswani2017attention} as the text encoder $\mathrm{CLIP_{txt}}(\cdot)$.
Given input texts, it converts these texts into word vectors through tokenization and a lookup procedure.
$\mathrm{CLIP_{txt}}(\cdot)$ outputs text features using these word vectors, and these features are mapped onto the same joint vision-language embedding space through $\ell_{2}$ normalization.
We also keep $\mathrm{CLIP_{txt}}(\cdot)$ frozen.

\begin{wrapfigure}{r}{0.5\textwidth}
  \vspace{-15.5mm}
  \begin{center}
    \includegraphics[width=0.5\textwidth]{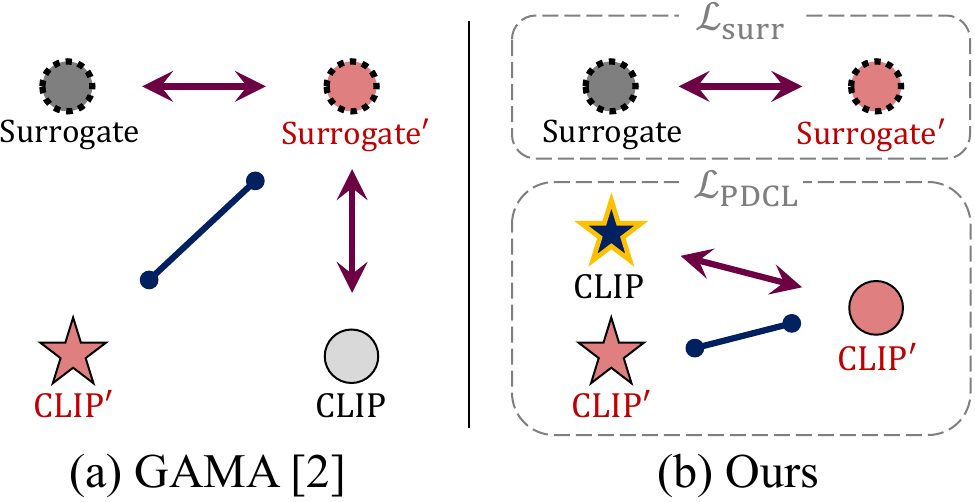}
  \end{center}
  \vspace{-6mm}
  \caption{\textbf{Loss design.} We separate the heterogeneous features extracted from each surrogate and CLIP model.
  {\raisebox{-0.575mm}{\LARGE{$\circ$}}} and \raisebox{-0.25mm}{{{\pdfliteral direct {2 Tr 0.25 w}{{\fontsize{10.6pt}{13pt}\selectfont \ding{73}}}\pdfliteral direct {0 Tr 0 w}}}} represent image and text features, respectively.
  \textcolor{red}{\textbf{Red}} denotes the adversarial features, and \textcolor{Goldenrod}{\textbf{Yellow}} denotes the features from GT label.}
  \vspace{-6.5mm}
  \label{fig:supp_comparison_gama}
\end{wrapfigure}

\subsection{Distinction from GAMA~\cite{aich2022gama}}
GAMA~\cite{aich2022gama} fundamentally differs from our work as it employs CLIP~\cite{CLIP} to attack \textit{multi-object scenes}, whereas our work emphasizes on enhancing transferable attacks by fully leveraging CLIP's capabilities and potential.
This section elaborates on our key contributions and highlights the distinctions from GAMA~\cite{aich2022gama}.

\noindent\textbf{Loss design.}\quad
GAMA~\cite{aich2022gama} mixes and contrasts heterogeneous features from two different representation spaces: surrogate and CLIP model.
We observed that blending of incompatible features extracted from models of different architectures and training methodologies could conflict, thus hindering optimization.
As described in their paper~\cite{aich2022gama}, this design constrains the mid-layer feature dimension of the surrogate model to match that of the CLIP embedding (i.e., 512).
To address this limitation and tackle the feature heterogeneity, we separate the surrogate and CLIP feature spaces as shown in Fig.~\ref{fig:supp_comparison_gama}.
In this approach, we can select any intermediate layer from the surrogate model without constraints, and leverage highly effective mid-layer features, as validated by BIA~\cite{zhang2022beyond}.
We provide layer ablation results in Fig.~\ref{fig:supp_layer_ablation}.
Additionally, our method leverages CLIP's text features from ground-truth (GT) class labels and CLIP's perturbed image features to provide effective loss gradients for training the generator.

\noindent\textbf{Prompt design.}\quad
GAMA~\cite{aich2022gama} employs a hand-crafted version of context words (\ie, ``a photo depicts''), whereas we investigate the effect of text prompts, as demonstrated in Table~5.
To fully unleash the power of text-driven features, we introduce a prior prompt learning stage to enhance the robustness of CLIP to distribution shifts.
To the best of our knowledge, we are the first to leverage prompt learning to enhance the generator-based transfer attack.

\noindent\textbf{Attack effectiveness.}\quad
Compared to GAMA~\cite{aich2022gama} using the same baseline, our method with enhanced attack loss design (\textbf{Ours}\textsuperscript{\textdagger}, $1.87\%p\uparrow$) and incorporating prompt learning (\textbf{Ours}, $4.65\%p\uparrow$) demonstrates significant improvements.

\setcounter{figure}{0}
\setcounter{table}{0}
\setcounter{algorithm}{0}
\setcounter{equation}{0}
\renewcommand{\thefigure}{B\arabic{figure}}
\renewcommand{\thetable}{B\arabic{table}}
\renewcommand{\thealgorithm}{B\arabic{algorithm}}
\renewcommand{\theequation}{B\arabic{equation}}
\renewcommand{\thesection}{B}

\section{Implementation Details}
\label{sec:supp_implementation_details}
Regarding the baseline framework, we follow the implementation of recent generative model-based attack methods~\cite{poursaeed2018generative, naseer2019cross, salzmann2021learning, zhang2022beyond, aich2022gama, yang2024facl} to establish a fair comparison.
The model architecture of the perturbation generator $G_\theta(\cdot)$ is composed of multiple down-sampling, residual, and up-sampling blocks.
The generator is tasked with crafting unbounded adversarial examples from clean input images.
Subsequently, the generated unbounded adversarial examples are constrained within a pre-defined perturbation budget of $\ell_{\infty} \leq 10$.
Then, pairs of adversarial images and their corresponding clean counterparts are fed into the surrogate model to perform an attack, simulating potential victim models.

\noindent\textbf{Computational specifics.}\quad
The training process takes approximately 33 hours on a single NVIDIA RTX A6000 GPU.
The S/W stack includes Python 3.7.15, PyTorch 1.8.0, torchvision 0.9.0, CUDA 11.1, and cuDNN 8.4.1.

\noindent\textbf{Dataset details.}\quad
As our primary objective is to produce highly transferable adversarial examples across different domains and models, we conduct experiments in rigorous black-box scenarios, specifically in \textit{cross-domain} and \textit{cross-model} settings.
Building upon insights from a recent study~\cite{zhang2022beyond}, which demonstrated effective attack transferability using a large-scale training dataset such as ImageNet-1K~\cite{imagenet}, we have also trained our perturbation generator on ImageNet-1K.
To assess the efficacy of the trained generator, we validate its performance by conducting evaluations on three additional datasets: CUB-201-2011~\cite{cub}, Stanford Cars~\cite{car}, and FGVC Aircraft~\cite{air}.
In the \textit{cross-domain} scenario, we gauge the effectiveness of our approach by evaluating it against unknown target domains and victim models that differ from the source domain and surrogate model.
In the \textit{cross-model} scenario, we evaluate our method against black-box models while maintaining a white-box domain configuration, specifically relying on ImageNet-1K.
Additionally, we incorporate four additional datasets to tackle the distribution-shifting scenario: ImageNet-V2~\cite{imagenet-v2}, ImageNet-Sketch~\cite{imagenet-sketch}, ImageNet-A~\cite{imagenet-a}, and ImageNet-R~\cite{imagenet-r}.
These datasets are employed during Phase 1 of prompt context optimization to validate the effectiveness of the trained $\texttt{Prompter}(\cdot)$.
We provide detailed information on the datasets used in our main experiments in Table~\ref{tab:supp_datasets}.

\begin{table}[!t]
\caption{Dataset details.}
\centering
\setlength{\tabcolsep}{4.5pt}
\resizebox{0.7\linewidth}{!}{
    \begin{tabular}{l|ccc}
        \toprule
        Dataset & $\#$ Class & $\#$ Train & $\#$ Validation \\
        \midrule
        \midrule
        ImageNet-1K~\cite{imagenet} & 1,000 & 1.28\;M & 50,000 \\
        CUB-200-2011~\cite{cub} & 200 & 5,994 & 5,794 \\
        Stanford Cars~\cite{car} & 196 & 8,144 & 8,041 \\
        FGVC Aircraft~\cite{air} &  100 & 6,667 & 3,333 \\
        \bottomrule                     
    \end{tabular}
    }
\label{tab:supp_datasets}
\end{table}
\begin{table}[!t]
    \caption{Improved robustness of CLIP~\cite{CLIP} to distribution shifts with $\texttt{Prompter}(\cdot)$.}
    \centering
    \setlength{\tabcolsep}{5pt}
    \renewcommand{\arraystretch}{1.2}
    \renewcommand{\aboverulesep}{0.1pt}
    \renewcommand{\belowrulesep}{0.1pt}
    \resizebox{0.85\linewidth}{!}{
    \begin{tabular}{l|c|cccc}
         \toprule
         Method & ImageNet-1K & -V2 & -Sketch & -A & -R \\
         \midrule
         \midrule
         Zero-shot CLIP~\cite{CLIP} & 66.7 & 60.9 & 46.1 & 47.8 & 74.0 \\
         \rowcolor{gray!9.0}w/ $\texttt{Prompter}(\cdot)$ & \textbf{71.9} & \textbf{64.2} & \textbf{46.3} & \textbf{48.9} & \textbf{74.6} \\
         \bottomrule
    \end{tabular}
    }
    \label{tab:supp_distribution_shifts}
\end{table}

\noindent\textbf{Prompter.}\quad
Prompt context training (Phase 1) is carried out prior to training of the perturbation generator (Phase 2).
This is because the trained $\texttt{Prompter}(\cdot)$ can extract more generalizable features from training on a large-scale dataset (\textit{i.e.}, ImageNet-1K~\cite{imagenet}), which contains diverse distribution shifts, thereby enhancing feature contrastive training in Phase 2.
Regarding the prompt learning, we follow the standard protocol~\cite{CoOp, CLIP} to guarantee the broad applicability of our method.
Specifically, we model the context words used in $\texttt{Prompter}(\cdot)$ using continuous learnable vectors as $[\mathbf{V}_{1}] [\mathbf{V}_{2}] \cdots [\mathbf{V}_{M}]$.
These vectors function as a unified class prefix, a strategy known to exhibit superior performance compared to a class-specific context~\cite{CoOp}.
The vectors are initialized with random values drawn from a zero-mean Gaussian distribution with a standard deviation of $0.02$, and each $\mathbf{V}_{i}$ has a word embedding size of 512, consistent with CLIP~\cite{CLIP}.
The number of context tokens $M$ is set to $16$.
For the hyperparameters, we follow the standard protocol outlined in \cite{CoOp}.
We use the Stochastic Gradient Descent (SGD) optimizer with a learning rate of $0.002$, which undergoes decay based on a cosine annealing scheduler.
We optimize the prompt context through a maximum of $50$ epochs using a 16-shot learning approach on ImageNet-1K~\cite{imagenet}.
In addition, we employ distribution-shifted variants of ImageNet, including -V2~\cite{imagenet-v2}, -Sketch~\cite{imagenet-sketch}, -A~\cite{imagenet-a}, and -R~\cite{imagenet-r}, to assess the robustness of the trained prompter.
As shown in Table~\ref{tab:supp_distribution_shifts}, our trained $\texttt{Prompter}(\cdot)$ improves the classification performance of the original CLIP model across various distribution scenarios.
In Phase 2, we leverage this generalized $\texttt{Prompter}(\cdot)$ to boost the prompt-driven feature contrastive training, facilitating more transferable attacks.

\noindent\textbf{Vision-language model.}\quad
We adopt CLIP~\cite{CLIP} as our vision-language model and employ the publicly available model.\footnote{\url{https://github.com/openai/CLIP}}
We utilize ViT-B/16~\cite{dosovitskiy2020vit} as the image encoder and the Transformer~\cite{vaswani2017attention} as the text encoder, due to the effective representation power of vision transformers (ViT)~\cite{dosovitskiy2020vit}.
This aligns with the empirical findings in GAMA~\cite{aich2022gama}, where ViT-B/16 is utilized and demonstrates superior performance compared to other backbones.

\noindent\textbf{Competitors.}\quad
For GAP~\cite{poursaeed2018generative}, we used their official code to train the generator.
For CDA~\cite{naseer2019cross}, we utilized their publicly available pre-trained models for evaluation.
We re-implemented LTP~\cite{salzmann2021learning} using the same generator model as BIA~\cite{zhang2022beyond}.
We adopted BIA~\cite{zhang2022beyond} as our baseline and leveraged their original codebase.
Regarding GAMA~\cite{aich2022gama}, we customized their official code to adapt it for training on ImageNet-1K~\cite{imagenet}.
We train all the baselines~\cite{poursaeed2018generative, naseer2019cross, salzmann2021learning, zhang2022beyond, aich2022gama} on the same ImageNet-1K dataset to ensure a fair comparison.

\setcounter{figure}{0}
\setcounter{table}{0}
\setcounter{algorithm}{0}
\setcounter{equation}{0}
\renewcommand{\thefigure}{C\arabic{figure}}
\renewcommand{\thetable}{C\arabic{table}}
\renewcommand{\thealgorithm}{C\arabic{algorithm}}
\renewcommand{\theequation}{C\arabic{equation}}
\renewcommand{\thesection}{C}

\section{More Quantitative Results}
\label{sec:supp_quantitative}
\begin{table}[!t]
\centering
\caption{\textbf{Multiple random runs with three different seeds.} We report the averaged top-1 classification accuracy after attacks with the standard deviation.}
\setlength{\tabcolsep}{3.5pt}
\renewcommand{\arraystretch}{1.2}
\renewcommand{\aboverulesep}{0.1pt}
\renewcommand{\belowrulesep}{0.1pt}
\resizebox{0.5\linewidth}{!}{
\begin{tabular}{c|cc}
\toprule
Method & \multicolumn{1}{c}{Cross-Domain} & \multicolumn{1}{c}{Cross-Model}  \\
\midrule
\midrule
Clean & 90.85 & 75.63 \\
\midrule
Run 1 & 43.91 & 38.60 \\
Run 2 & 44.24 & 39.51 \\
Run 3 & 43.38 & 40.06 \\
\midrule
\rowcolor[gray]{0.95}Mean \scriptsize{{$\pm$} Std.}  & 43.84 \scriptsize{$\pm$ 0.43} & 39.39 \scriptsize{$\pm$ 0.60} \\
\bottomrule
\end{tabular}
}
\label{tab:supp_multiple_random_runs}
\end{table}
\noindent\textbf{Multiple random runs.}\quad
We conducted three separate runs with different random seeds to ensure the reproducibility of our proposed method, and the results are shown in Table~\ref{tab:supp_multiple_random_runs}.
Our method consistently demonstrates reliable performance across multiple random runs.

\begin{table}[!t]
    \centering
    \caption{\textbf{Comparison with iterative attacks.} We report the attack success rate (ASR) on a balanced subset of ImageNet-1K~\cite{imagenet}.    \textbf{Ours}\textsuperscript{\textdagger} denotes the results using a hand-crafted context prompt (\textit{i.e.}, ``a photo of a [class]'').
    \textbf{Best} and \underline{second best}.
    }
    \renewcommand{\arraystretch}{1.2}
    \renewcommand{\aboverulesep}{0.1pt}
    \renewcommand{\belowrulesep}{0.1pt}
    \setlength{\tabcolsep}{6pt}
    \resizebox{0.85\linewidth}{!}{
    \begin{tabular}{c|cccc|cc}
    \toprule
    
    Method & DI~\cite{DI} & TI~\cite{TI} & SI~\cite{SI} & Admix~\cite{Admix} & \cellcolor[gray]{0.95}\textbf{Ours}\textsuperscript{\textdagger} & \cellcolor[gray]{0.95}\textbf{Ours} \\
    \midrule
    \midrule

    ASR(\%)$\uparrow$ & 28.08 & 33.92 & 37.44 & 30.80 & \cellcolor[gray]{0.95}\underline{50.68} & \cellcolor[gray]{0.95}\textbf{54.78} \\
    
    \bottomrule
    \end{tabular}
    }
    \label{tab:supp_iterative}
\end{table}

\noindent\textbf{Comparison with iterative attacks.}\quad
Due to the significant computational burden of gradient-based iterative attacks, we evaluate the attack success rates (ASR (\%)$\uparrow$) using a smaller and balanced subset of ImageNet-1K~\cite{imagenet}, as shown in Table~\ref{tab:supp_iterative}.
To create the smaller subset, we randomly selected 5,000 images (5 per class) from the validation set, ensuring that all of these images are correctly classified.
Regarding the hyperparameters, we follow the standard settings deployed in the original paper.
We set the step size $\alpha=4$, and the number of iterations $T=100$ for all the iterative methods.
For DI~\cite{DI}, we set the decay factor $\mu=1.0$ and the transformation probability $p=0.7$.
All methods have been trained using VGG-16~\cite{vgg} and evaluated on Inception-v3 (Inc-v3)~\cite{inc-v3} with $\ell_{\infty} \leq 10$.
The results demonstrate the superiority of our generative attack method in terms of attack transferability.
Notably, \textbf{Ours}\textsuperscript{\textdagger} achieves better results even without prompt learning, indicating that our generative attack method leveraging CLIP-driven feature guidance is highly effective on its own.

\begin{table}[!t]
    \centering
    \caption{\textbf{Attack effectiveness against defense methods.} Our method retains superior attack effectiveness compared to other baselines.
    \textbf{Best} and \underline{second best}.
    }
    \setlength{\tabcolsep}{4pt}
    \renewcommand{\arraystretch}{1.2}
    \renewcommand{\aboverulesep}{0.1pt}
    \renewcommand{\belowrulesep}{0.1pt}
    \resizebox{0.8\linewidth}{!}{
    \begin{tabular}{c|ccc|c}
    \toprule
    
    Method & Adv. Inc-v3 & Adv. IncRes-v2$_{\mathrm{ens}}$ & JPEG~($75\%$) & \;AVG.\; \\
    \midrule
    \midrule
    
    Clean & 76.34 & 78.64 & 74.68 & 76.58  \\
    \midrule
    
    BIA~\cite{zhang2022beyond} & 68.52 & \underline{75.11} & 63.77 & 69.13 \\

    GAMA~\cite{aich2022gama} & \underline{66.72} & 75.52 & \underline{59.60} & \underline{67.28} \\
    \midrule
    
    \rowcolor[gray]{0.95}\textbf{Ours} & \textbf{66.29} & \textbf{75.10} &  \textbf{57.92} & \textbf{66.44} \\
    
    \bottomrule
    \end{tabular}
    }
    \label{tab:supp_defense}
\end{table}

\noindent\textbf{Attack effectiveness against defense methods.}\quad
In Table~\ref{tab:supp_defense}, we conduct additional evaluations to assess the effectiveness of our method against defense methods, including adversarially trained models (Adv. Inc-v3, Adv. IncRes-v2$_{\mathrm{ens}}$~\cite{wang2020towards}), and an input processing method (JPEG~\cite{JPEG}).
We train our perturbation generator using VGG-16~\cite{vgg} and report the average top-1 classification accuracy after attacks, where lower values indicate better performance.
For JPEG defense, we applied a compression rate of 75\%, and the victim model used for evaluation is Inc-v3~\cite{inc-v3}.
Our method consistently demonstrates superior attack effectiveness against defense methods.

\begin{table}[!t]
    \caption{\textbf{Cross-model attack effectiveness w.r.t. perturbation budget.}
    Given the same test-time perturbation budget, our method consistently demonstrates superior attack effectiveness.
    \textbf{Best} and \underline{second best}.}
    \centering
    \setlength{\tabcolsep}{5pt}
    \renewcommand{\arraystretch}{1.2}
    \renewcommand{\aboverulesep}{0.1pt}
    \renewcommand{\belowrulesep}{0.1pt}
    \resizebox{0.65\linewidth}{!}{
    \begin{tabular}{c|ccccc}
         \toprule
         $\ell_{\infty} \leq$ & 6 & 7 & 8 & 9 & 10 \\
         \midrule
         \midrule
         
         BIA~\cite{zhang2022beyond} & 63.67 & 58.48 & 51.85 & 47.82 & 42.93\\
         GAMA~\cite{aich2022gama} & \underline{61.62} & \underline{56.20} & \underline{50.74} & \underline{45.57} & \underline{40.91} \\
         \midrule

         \rowcolor{gray!9.0}\textbf{Ours} & \textbf{60.41} & \textbf{54.64} & \textbf{48.84} & \textbf{43.47} & \textbf{38.60} \\
         
         \bottomrule
         
    \end{tabular}
    }
    \label{tab:perturbation_budget_cross_model}
\end{table}

\noindent\textbf{Cross-model transferability w.r.t. perturbation budget.}\quad
We evaluated the top-1 attack accuracy with various test-time perturbation budgets, as shown in Table~\ref{tab:perturbation_budget_cross_model}.
Our method also proves effective transferability across various model architectures, consistently outperforming the baseline scores.
In other words, our method can achieve higher attack effectiveness with lower perturbation power and better image quality, offering significant advantages for deployment.

\begin{figure}[!t]
\begin{minipage}[b]{0.48\linewidth}
\centering
\includegraphics[width=\textwidth]{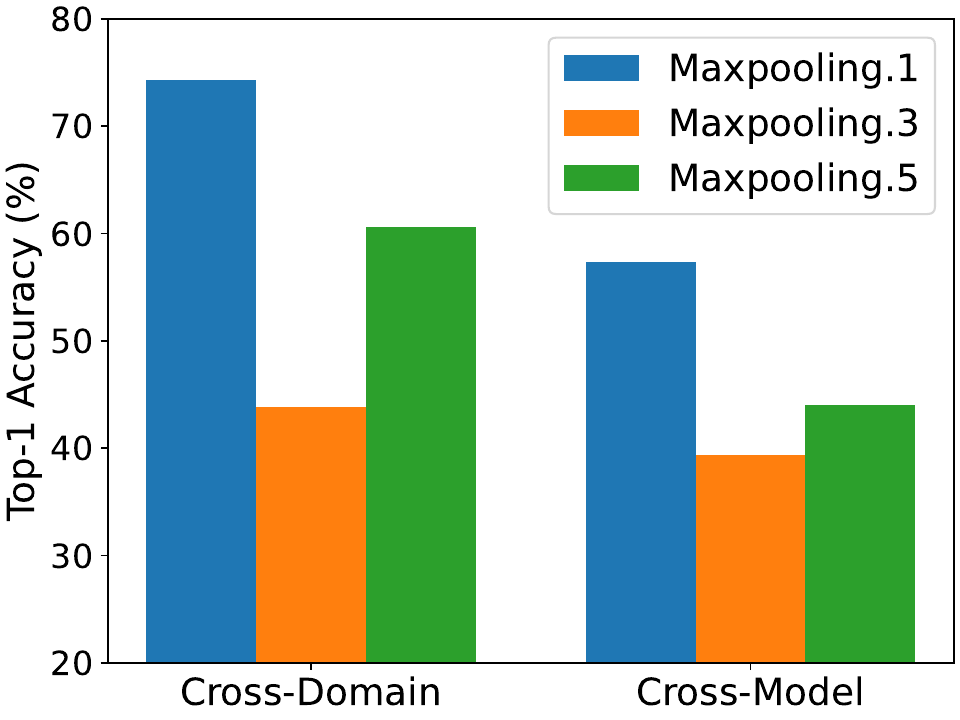}
\caption{\textbf{Selection of mid-layer features.} Varying the selection of mid-layer from surrogate model (VGG-16~\cite{vgg}), we report the averaged top-1 accuracy after attacks (the lower, the better) on both cross-domain and cross-model settings.}
\label{fig:supp_layer_ablation}
\end{minipage}
\hspace{0.02\linewidth}
\begin{minipage}[b]{0.48\linewidth}
\centering
\includegraphics[width=\textwidth]{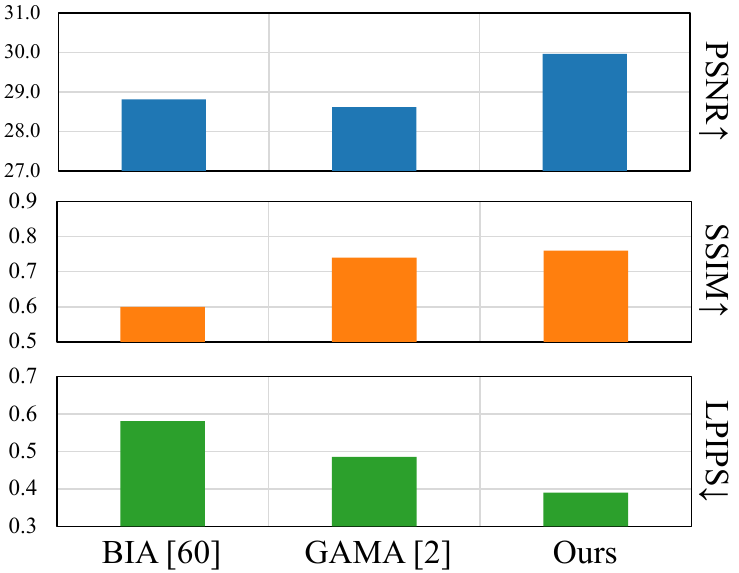}
\caption{\textbf{Analysis on perceptual image quality.} Across the three perceptual image quality metrics (\ie, PSNR, SSIM, and LPIPS), our method demonstrates superior image quality scores while retaining superior attack effectiveness.}
\label{fig:supp_image_quality}
\end{minipage}
\end{figure}

\noindent\textbf{Selection of mid-layer features.}\quad
In Fig.~\ref{fig:supp_layer_ablation}, we examine the influence of selecting various intermediate layers from the surrogate model of VGG-16~\cite{vgg} on the transferability of the generated adversarial examples.
Each result represents the averaged top-1 accuracy after attacks across three different random seeds.
Consistent with BIA~\cite{zhang2022beyond}, which leverages middle-level layers for enhanced transferability, our method also demonstrates superior attack transferability when selecting the mid-level layer of \textit{Maxpool.}3 for effective feature extraction.

\noindent\textbf{Analysis on perceptual image quality.}\quad
As shown in Fig.~\ref{fig:supp_image_quality}, we conducted  quantitative evaluation on the perceptual image quality of the generated adversarial examples.
Across the three perceptual image quality metrics (\textit{i.e.}, PSNR, SSIM, and LPIPS), our method shows consistently better scores than prior methods.
Remarkably, our method enhances perceptual image quality while effectively improving the attack performance.
We posit that our prompt-driven feature guidance in the joint vision-language embedding space facilitates more realistic guidance towards natural images.
The relatively degraded image quality of BIA~\cite{zhang2022beyond} could be attributed to the generator training solely induced by perturbing critical image features from the surrogate model.
This underscores another advantage of incorporating the CLIP model into the attack framework, in addition to enhancing the attack effectiveness.

\noindent\textbf{Training against other surrogate model.}\quad
In all of our main experiments, we employ the ImageNet-1K~\cite{imagenet} pre-trained VGG-16~\cite{vgg} as our surrogate model.
We additionally trained the generator against Dense-169~\cite{densenet} and present evaluation results in both cross-domain and cross-model settings, as shown in Table~\ref{tab:supp_surrogate_dense169_cross_domain} and Table~\ref{tab:supp_surrogate_dense169_cross_model}, respectively.
Our method consistently enhances attack transferability across different settings, highlighting the versatility and generalizability of our approach.

\begin{table*}[!t]
\caption{\textbf{Cross-domain evaluation results.} The perturbation generator is trained on ImageNet-1K~\cite{imagenet} with Dense-169~\cite{densenet} as the surrogate model and evaluated on black-box domains with models.
We compare the top-1 classification accuracy after attacks ($\downarrow$ is better) with the perturbation budget of $\ell_\infty \leq 10$.
\textbf{Best} and \underline{second best}.}
\setlength{\tabcolsep}{1.1pt}
\renewcommand{\arraystretch}{1.2}
\renewcommand{\aboverulesep}{0.1pt}
\renewcommand{\belowrulesep}{0.1pt}
\centering
\resizebox{\linewidth}{!}{
    \begin{tabular}{ccccccccccccc}
        \toprule
        \multirow{2}{*}{Method} & \multicolumn{3}{c}{CUB-200-2011} && \multicolumn{3}{c}{Stanford Cars} && \multicolumn{3}{c}{FGVC Aircraft} & \multirow{2}{*}{AVG.} \\
        \cmidrule{2-12}
         
        & Res-50 & SENet154 & SE-Res101 && Res-50 & SENet154 & SE-Res101 &&  Res-50 & SENet154 & SE-Res101 \\
        \midrule
        \midrule
        
        Clean & 87.33 & 86.81 & 86.59 && 94.25 & 93.35 & 92.96 && 92.14 & 92.05 & 91.84 & 90.81 \\
        \midrule
        
        GAP~\cite{poursaeed2018generative} & 60.87 & 72.39 & 68.17 && 77.63 & 83.72 & 84.84 && 75.46 & 80.02 & 72.64 & 75.08\\
        
        CDA~\cite{naseer2019cross} & 52.92 & 60.96 & 57.04 && 53.64 & 73.66 & 75.51 && 62.23 & 61.42 & 59.83 &61.91\\
        
        LTP~\cite{salzmann2021learning} & \underline{19.97} & 34.09 & 45.48 && \textbf{4.81}  & 47.61 & \underline{46.05} && \underline{5.19}  & \underline{19.71} & \underline{26.16} & 27.67\\
        
        BIA~\cite{zhang2022beyond} & 21.79 & 29.29 & 39.13 && 9.58 & 44.46 & 49.06 && 8.04 & 27.84 & 33.87 & 29.23 \\

        GAMA~\cite{aich2022gama} & 21.02 & \textbf{25.30} & \underline{38.21} && 8.68 & \underline{37.55} & 51.20 && 7.14 & 22.59 & 35.43 & \underline{27.46} \\
        \midrule
        
        \rowcolor{gray!9.0}\textbf{Ours} & \textbf{10.53}  & \underline{28.27}  & 	\textbf{37.64}  && 	\underline{6.67}  & 	\textbf{35.46}  & 	\textbf{31.92}  && 	\textbf{3.27}  & 	\textbf{12.00}  & 	\textbf{16.47}  & 	\textbf{20.25}  \\

        \bottomrule
    \end{tabular}
}
\label{tab:supp_surrogate_dense169_cross_domain}
\end{table*}
\begin{table*}[!t]
\caption{\textbf{Cross-model evaluation results.}
The perturbation generator is trained on ImageNet-1K~\cite{imagenet} with Dense-169~\cite{densenet} as the surrogate model and evaluated on black-box models.
We compare the top-1 classification accuracy after attacks ($\downarrow$ is better) with the perturbation budget of $\ell_\infty \leq 10$.
\textbf{Best} and \underline{second best}.}
\setlength{\tabcolsep}{2.5pt}
\renewcommand{\arraystretch}{1.2}
\renewcommand{\aboverulesep}{0.1pt}
\renewcommand{\belowrulesep}{0.1pt}
\centering
\resizebox{\linewidth}{!}{
    \begin{tabular}{cccccccccc}
        \toprule 
        Method & 
        VGG-16 & VGG-19 &
        Res-50 & Res-152 &
        Inc-v3 & MNasNet &
        ViT-B/16 & ViT-L/16 &
        AVG. \\
        \midrule
        \midrule
        
        Clean 
        & 70.14 & 70.95
        & 74.61 & 77.34
        & 76.19 & 66.49
        & 79.56 & 80.86
        & 74.52 \\
        \midrule
    
        GAP~\cite{poursaeed2018generative}
        & 39.11 & 39.62
        &50.72 & 58.33
        & 48.08 & 44.63
        & 72.28 & 74.71
        & 53.44 \\
          
        CDA~\cite{naseer2019cross} 
        & 7.26 & 7.91
        & 6.46 & 15.56
        & 43.78 & \textbf{21.89} &
        \textbf{54.23} & 68.14 &
        28.15 \\
        
        LTP~\cite{salzmann2021learning} 
        & 5.93 & 7.52
        & 6.34 & 10.73
        & 40.92 & 36.94
        & 64.49 & 73.01
        & 30.74 \\
        
        BIA~\cite{zhang2022beyond} 
        & 4.76 & 7.15
        & 6.97 & 13.83
        & 38.58 & 24.32
        & 57.90 & 67.78
        & 27.66 \\

        GAMA~\cite{aich2022gama} 
        & 2.87 & \textbf{4.96}
        & \textbf{4.24} & \underline{10.50}
        & \underline{32.11} & 24.23
        & \underline{55.19} & \textbf{66.48}
        & \underline{25.07} \\ 
        \midrule

        \rowcolor{gray!9.0}\textbf{Ours} 
        & \textbf{2.62} & \underline{5.58}
        & \underline{5.67} & \textbf{8.45}
        & \textbf{21.33} & \underline{22.59}
        & 55.34 & \underline{67.31}
        & \textbf{23.61} \\
        
        \bottomrule
         
    \end{tabular}
}
\label{tab:supp_surrogate_dense169_cross_model}
\end{table*}

\setcounter{figure}{0}
\setcounter{table}{0}
\setcounter{algorithm}{0}
\setcounter{equation}{0}
\renewcommand{\thefigure}{D\arabic{figure}}
\renewcommand{\thetable}{D\arabic{table}}
\renewcommand{\thealgorithm}{D\arabic{algorithm}}
\renewcommand{\theequation}{D\arabic{equation}}
\renewcommand{\thesection}{D}

\section{More Qualitative Results}
\label{sec:supp_qualitative}
\begin{figure}
    \centering
    \includegraphics[width=\linewidth]{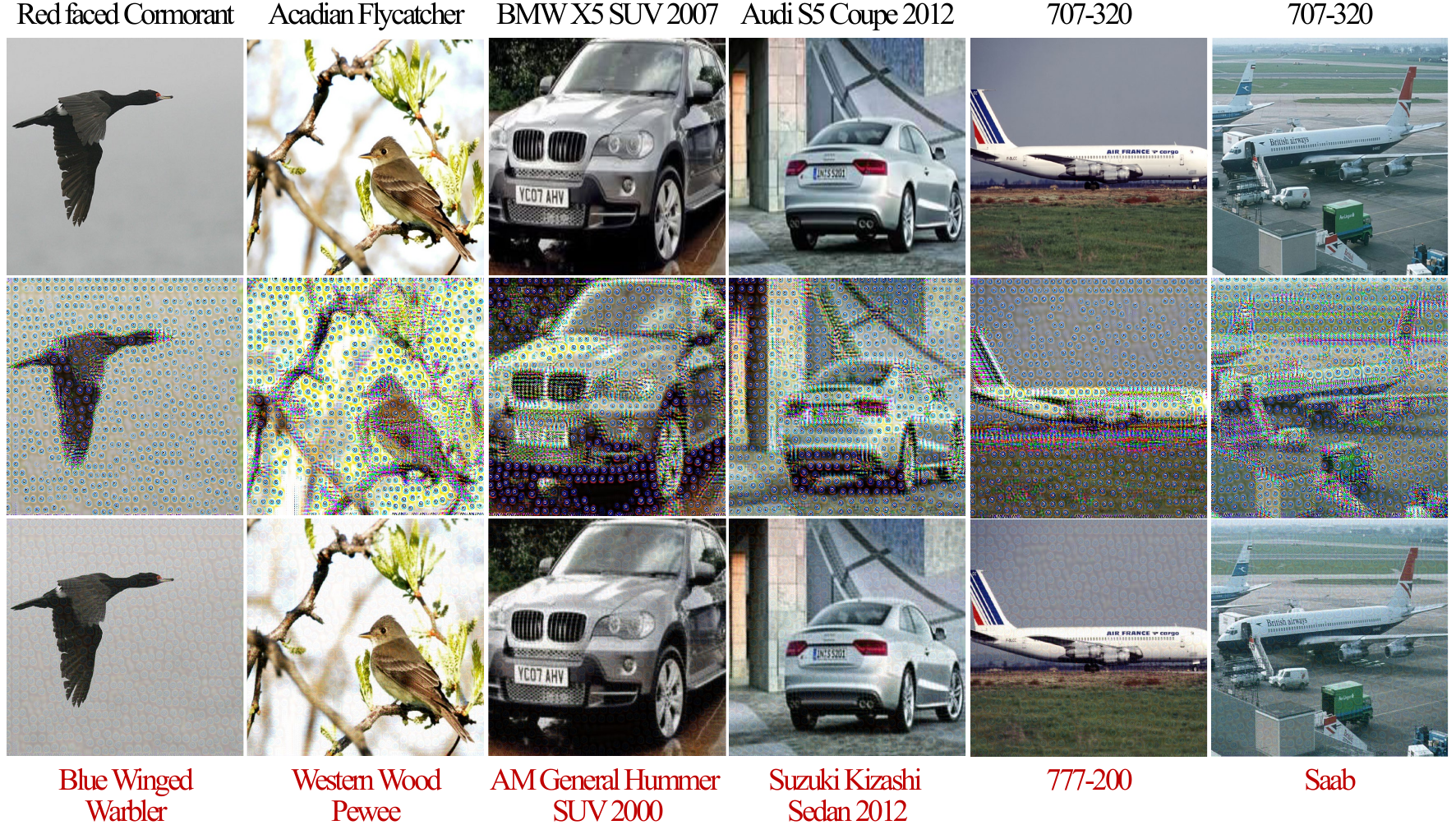}
    \caption{\textbf{Qualitative results on CUB-200-2011, Stanford Cars, and FGVC Aircraft.} Clean images (\textit{row 1}), unbounded adversarial images (\textit{row 2}), and bounded ($\ell_{\infty}\le 10$) adversarial images (\textit{row 3}; actual inputs to the classifier) are shown.
    The ground truth and each mis-predicted class label are shown on the \textit{top} and \textit{bottom}.}
    \label{fig:supp_qualitative_domain}
\end{figure}

\begin{figure}
\centering
    \includegraphics[width=\linewidth]{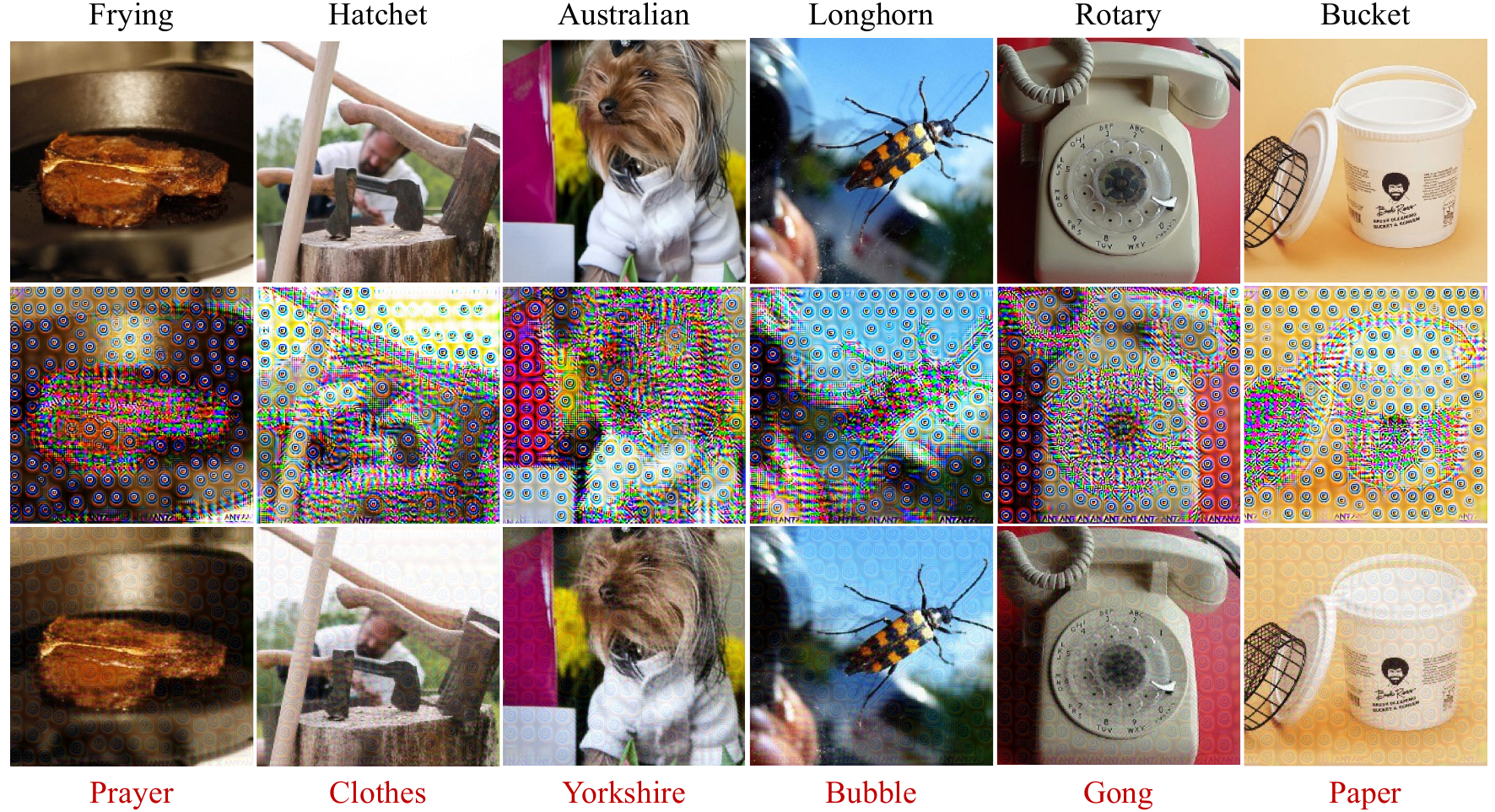}
    \caption{\textbf{Qualitative results on ImageNet-1K}. Clean images (\textit{row 1}), unbounded adversarial images (\textit{row 2}), and bounded ($\ell_{\infty}\le 10$) adversarial images (\textit{row 3}; actual inputs to the classifier) are shown.
    The ground truth and each mis-predicted class label are shown on the \textit{top} and \textit{bottom}.}
    \label{fig:supp_qualitative_imagenet}
\end{figure}

\begin{figure}
    \centering
    \includegraphics[width=\linewidth]{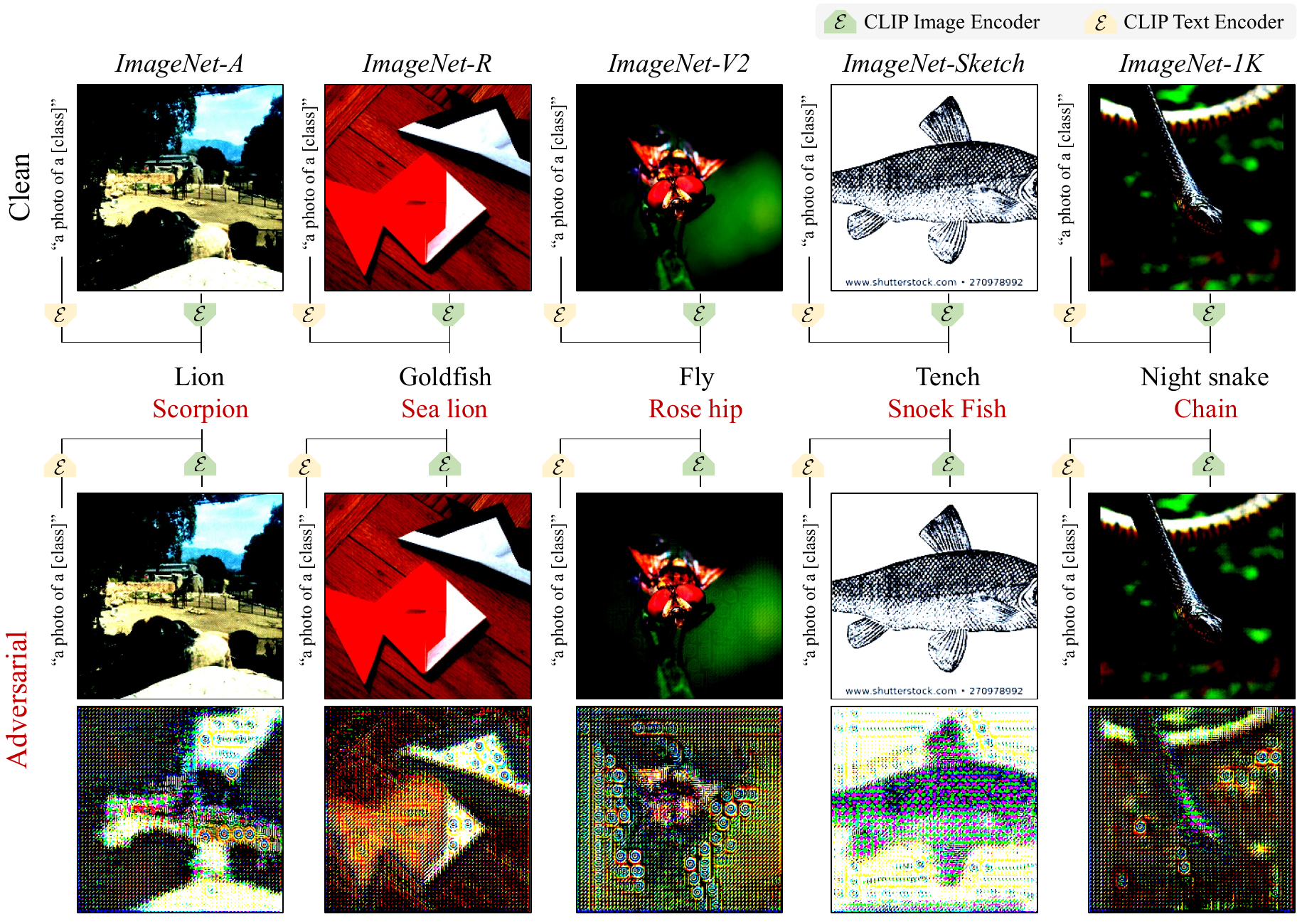} 
    \caption{\textbf{Qualitative results on ImageNet-1K and its distribution-shifted variants.} From top to bottom: Clean images, bounded ($\ell_{\infty}\le 10$) adversarial images, and unbounded adversarial images, respectively.
    In the middle, zero-shot CLIP-predicted class labels are displayed for both clean and adversarial image inputs.
    Our method effectively induces the zero-shot CLIP model to misclassify images as incorrect labels, even when faced with various distribution shifts.
    For the inference, we employ the text prompt ``a photo of a [class]'', following the common approach in CLIP~\cite{CLIP}.}
    \label{fig:supp_qualitative_zsclip}
\end{figure}

Additional adversarial image samples across diverse domains, including CUB-200-2011~\cite{cub}, Stanford Cars~\cite{car}, and FGVC Aircraft~\cite{air}, are depicted in Fig.~\ref{fig:supp_qualitative_domain}.
While unbounded adversarial images in \textit{row 2} may appear to significantly disrupt the classifier's predictions, the actual inputs to the classifier are those in \textit{row 3}, which have been clipped according to the perturbation budget of $\ell_{\infty} \leq 10$.
We further visualized adversarial image samples on ImageNet-1K~\cite{imagenet} in Fig.~\ref{fig:supp_qualitative_imagenet}, where our method successfully induces misclassification in unknown victim models.
Fig.~\ref{fig:supp_qualitative_zsclip} illustrates zero-shot CLIP classification results on ImageNet-1K and its distribution-shifted variants such as ImageNet-A~\cite{imagenet-a}, ImageNet-R~\cite{imagenet-r}, ImageNet-V2~\cite{imagenet-v2}, and ImageNet-Sketch~\cite{imagenet-sketch}.
By incorporating CLIP-driven feature guidance into the generative attack framework, our method can also disrupt CLIP's zero-shot predictions as well.

\end{document}